\documentclass[final,3p,times,twocolumn]{elsarticle}
\usepackage{hyperref}
\usepackage{amssymb}
\usepackage{booktabs} % 
\usepackage{amsmath}
\usepackage{cases}
\usepackage{graphicx}
\usepackage{subcaption}
\usepackage{algorithm,float}
\usepackage[noend]{algorithmic}
\usepackage{optidef}
\usepackage{ragged2e} 
\usepackage{makecell, multirow, tabularx}
\usepackage[table]{xcolor}
\usepackage{threeparttable} %
\usepackage{bm}
\usepackage{dblfloatfix}

\newtheorem{theorem}{Theorem}

\newtheorem{definition}[theorem]{Definition} 


\makeatletter
\newif\ifrevised
\revisedtrue        % ← change to \revisedfalse to close all color

\ifrevised
\newcommand{\revmwl}[1]{{\color{black}#1}}

\else
\fi

\journal{}

\begin{document}

\begin{frontmatter}

  \title{An Overtaking Trajectory Planning Framework Based on Spatio-temporal Topology and Reachable Set Analysis Ensuring Time Efficiency}
  
  \author[author1]{Wule Mao} %% Author name
  \author[author1]{Zhouheng Li}
  \author[author2]{Entao Sun}
  \author[author1]{Lei Xie\corref{corresponding}}
  \ead{leix@iipc.zju.edu.cn}
  \author[author1]{Hongye Su}
  \address[author1]{State Key Laboratory of Industrial Control Technology, Zhejiang University, Hangzhou, 310027, China}
  \address[author2]{Shanghai STEP Electric Corporation, Shanghai, 201802, China}
  \cortext[corresponding]{Corresponding author}
 
\begin{abstract}
\revmwl{Generating overtaking trajectories in high-speed scenarios is typically addressed through hierarchical planning, which often suffers from local optima due to single initial solutions and low computational efficiency during numerical optimization. To overcome these limitations, this paper proposes a \textbf{S}patio-temporal topology and \textbf{R}eachable set analysis enhanced \textbf{O}vertaking trajectory \textbf{P}lanning framework (SROP). Specifically, by introducing topological classes to represent distinct overtaking behaviors, the upper-layer planner performs a spatio-temporal search to extract diverse initial paths, effectively preventing local optima. Subsequently, a lower-layer planner conducts parallel trajectory evaluation using reachable sets, which decouples vehicle kinematic constraints from the optimization process to ensure feasibility and significantly accelerate computation. Numerical experiments demonstrate that SROP improves trajectory smoothness by 66.8\% and reduces computation time by 62.9\% compared to state-of-the-art methods. Furthermore, by seamlessly integrating the method into the F1TENTH autonomous racing simulation platform, a 100-lap sensitivity analysis demonstrates high overtaking success rates in challenging scenarios, thereby validating its practical utility, real-time efficiency, and robustness.}
\end{abstract}
  %% Keywords
  \begin{keyword}
    overtaking \sep topologies \sep spatio-temporal \sep reachable set
  \end{keyword}

\end{frontmatter}

\section{Introduction}
\label{introduction}
\revmwl{Autonomous driving technology has profoundly reshaped modern mobility and commercial logistics. As these systems are increasingly deployed in diverse, highly dynamic scenarios ranging from industrial automation and highway transportation to urban traffic and competitive racing~\cite{li6127037evo}, executing safe and efficient overtaking maneuvers has emerged as a universal requirement \cite{zhao2024autonomous}.  Consequently, trajectory planning plays a pivotal role by resolving the intricate spatial and temporal constraints of the overtaking action. An optimal planner enables a safe, efficient, and kinematically smooth execution~\cite{li2024rapid}, thereby elevating ride quality and ensuring physical limits are respected \cite{sun2025generalizing}.}

\revmwl{Trajectory planning typically employs a two-layer hierarchical framework \cite{hu2025survey}}. The upper layer uses heuristic methods to provide reference trajectories for the lower layer, which relies on optimization-based methods to generate trajectories for vehicle control. The upper layer determines a collision-free initial path, primarily addressing which side the vehicle should pass on. Meanwhile, the lower layer refines this initial solution through numerical optimization, ensuring it meets more complex constraints and achieves higher performance requirements. A critical insight of this method is dividing the entire configuration space into a series of non-convex subspaces. The upper-layer search aims to provide the lower-layer trajectory planning with a good initial solution and a non-convex subspace, thereby enhancing the efficiency of trajectory planning. Upper-layer trajectory planning methods generally include sampling and search techniques, which can quickly generate a collision-free initial trajectory. However, these methods have two significant drawbacks. \revmwl{First, distinct overtaking behaviors inherently require trajectories from different topological classes, yet traditional search or sampling techniques fail to provide diverse initial guesses that capture this topological variety~\cite{dixit2018trajectory,tian2025risk}.} \revmwl{Second, obstacles partition the environment into a non-convex feasible space, where continuous optimization algorithms typically converge only to the local optimum of their initial topology. Providing a single reference trajectory discards valuable alternative topologies, severely shrinking the solution space and often leaving lower-layer planners trapped in poor local minima}~\cite{rosmann2017integrated,ye2023std}.

\revmwl{Topologically distinct trajectories belong to different homotopy classes, each forming an isolated basin of attraction that represents a unique local optimum in the non-convex free space \cite{bhattacharya2012topological}}. To address these issues, this paper proposes an efficient spatio-temporal topological search method. This approach generates a series of collision-free initial paths from different topological classes through graph search. These initial paths act as a skeleton, providing reference values for detailed trajectory optimization at the lower layer. The critical insight is that different topological classes expand the solution space for lower-layer planning. This method can simultaneously consider multiple topological classes, each representing a different non-convex subspace. The advantage is that selecting the best local optimum from these subspaces as the final planning result enhances the solution quality.

The lower-layer trajectory planning method is used to generate a feasible trajectory for a vehicle based on the path identified by the upper layer. This planning method frames the entire trajectory generation problem as a constrained optimization model, which is then solved using numerical methods. \revmwl{It enforces the vehicle model as a hard constraint to strictly satisfy kinematic constraints  \cite{tang2025path}}. However, a significant drawback of these methods is that the constraints imposed by the vehicle model can limit the solution space, resulting in a complex and inefficient solving process, or even in cases where no solution exists. \revmwl{Consequently, this may lead to aborted overtaking maneuvers or severely compromise driving safety.}

To address this issue, we propose a trajectory generation method based on the reachable set of vehicles, efficiently producing trajectories while ensuring their kinematic feasibility. \revmwl{Theoretically, restricting trajectories within the forward reachable set inherently satisfies kinematic constraints, as it is computed subject to the vehicle's differential equations and input bounds \cite{althoff2014online}. Bounding state evolution within this Hamilton-Jacobi reachable tube implicitly enforces nonlinear constraints and precludes unexecutable motions \cite{bansal2017hamilton}. Consequently, leveraging these sets naturally enforces dynamic validity, eliminating the need for post-hoc trajectory smoothing \cite{ding2019safe}.} Specifically, it uses the initial paths provided by the upper-layer search as the reference trajectories. Multiple initial solutions from distinct topologies help prevent getting stuck in local optima, thereby enhancing the overall solution quality. Then, it generates a cluster of candidate trajectories in parallel. Finally, it evaluates the kinematic feasibility of these trajectories using the reachable set method and selects the optimal one as the overtaking trajectory. The key insight of this method is that it decouples trajectory generation from the vehicle model constraints by leveraging the reachable set method. The advantage of this approach is that it avoids the need to solve massive optimization problems, thereby enhancing the efficiency of trajectory optimization while ensuring the kinematic feasibility of the planned trajectories.
\subsection{Motivations}
\label{Motivations}
Hierarchical methods commonly used for planning overtaking trajectories encounter two main issues that need to be addressed:
\begin{itemize}
  \item[(1)] Search-based or sampling-based approaches in the upper layer often fail to produce trajectories that reflect diverse overtaking behaviors, which may lead to the planner getting trapped in local optima.
  \item[(2)] Optimization-based trajectory generation methods in the lower layer face challenges in balancing the efficiency and kinematic feasibility of the generated trajectories.
\end{itemize}

To tackle the issues mentioned, this paper introduces an overtaking trajectory planning framework based on spatio-temporal topology and reachable set analysis to generate an optimal trajectory that enforces kinematic feasibility efficiently. The upper layer of the trajectory planning utilizes a spatiotemporal topological search method through graph search, while the lower layer employs a reachable set-based parallel generation method to produce kinematically feasible trajectories. \revmwl{Furthermore, to bridge the gap between theoretical design and practical application, the proposed framework is seamlessly integrated into the F1TENTH autonomous racing simulation platform \cite{o2020f1tenth}, effectively validating its real-time robustness and safety in highly competitive scenarios.} The key innovations are as follows:
\subsection{Contributions}
\label{Contributions}
The contributions of this paper are summarized as follows:
\begin{itemize}
  \item [(1)]
        \textbf{An efficient spatio-temporal topological search method is proposed to avoid getting trapped in local optima.} This method identifies initial solutions from different topological classes to improve the trajectory quality. Using this method instead of a single initial solution reduces the trajectory tracking error by \textbf{19.6\%} and improves the average smoothness by \textbf{20.6\%}, demonstrating that the proposed method can avoid getting trapped in local optima, thereby improving the solution quality.
  \item [(2)]
        \textbf{The reachable set and parallel computation are introduced for rapid trajectory planning while ensuring kinematic feasibility.} While generating multiple candidate overtaking trajectories, the reachable set is used to perform selection to ensure kinematic feasibility. It provides time efficiency through parallel computation. The controller based on a pure pursuit algorithm validated the kinematic feasibility of the trajectory. Compared to the initial trajectory, the tracking error decreased by \textbf{40.3\%}, confirming the effectiveness of the proposed method in ensuring kinematic feasibility.
  \item [(3)]
        \textbf{An overtaking trajectory planning framework is proposed to ensure time efficiency and trajectory kinematic feasibility}.  Compared to state-of-the-art methods, this method demonstrates a \textbf{66.8\%} improvement in trajectory smoothness and a \textbf{62.9\%} reduction in computation time. This validates the effectiveness of the method in improving trajectory quality and planning efficiency.
\end{itemize}
The remaining paper is organized as follows: Section \ref{Related_work} reviews related work. Section \ref{preliminaries} describes relevant notations of reachable set analysis. Section \ref{sec_STHPTP} introduces the details of the overtaking trajectory planning framework based on spatiotemporal topology and reachable set analysis. Section \ref{Results} presents simulation results, and Section \ref{Conclusion} concludes the paper.

\section{Related Work}
\label{Related_work}
Overtaking trajectory planning in a road involves finding a path for the vehicle between a given start and goal that meets specific constraints and overtakes other vehicles. These constraints include safety (no collisions), minimum time, and minimum energy consumption. The hierarchical planning method is a typical approach to solving the above trajectory planning problem. The main idea of such methods is to obtain a geometric path in the configuration space based on the shortest path planning and then perform time parameterization or trajectory planning for this geometric path. An initial solution is provided by a search algorithm or dynamic programming algorithm, and then this initial solution is refined through nonlinear optimization in \cite{fan2018baidu,lim2018hierarchical}. A hierarchical method in \cite{xin2021enable} includes the front-end trajectory search and back-end optimization based on the convex feasible set algorithm proposed by \cite{liu2018convex}. A hierarchical method introduced by \cite{li2020road} first searches for an initial path in the spatiotemporal configuration space. Then it transforms the horizontal and vertical optimization problems into a quadratic programming problem to obtain a trajectory that is the shortest in length and the smoothest.

\begin{figure}[!t]%% placement specifier
  %% Use \includegraphics command to insert graphic files. Place graphics files in 
  %% working directory.
  \centering%% For centre alignment of image.
  \includegraphics[width=0.46\textwidth]{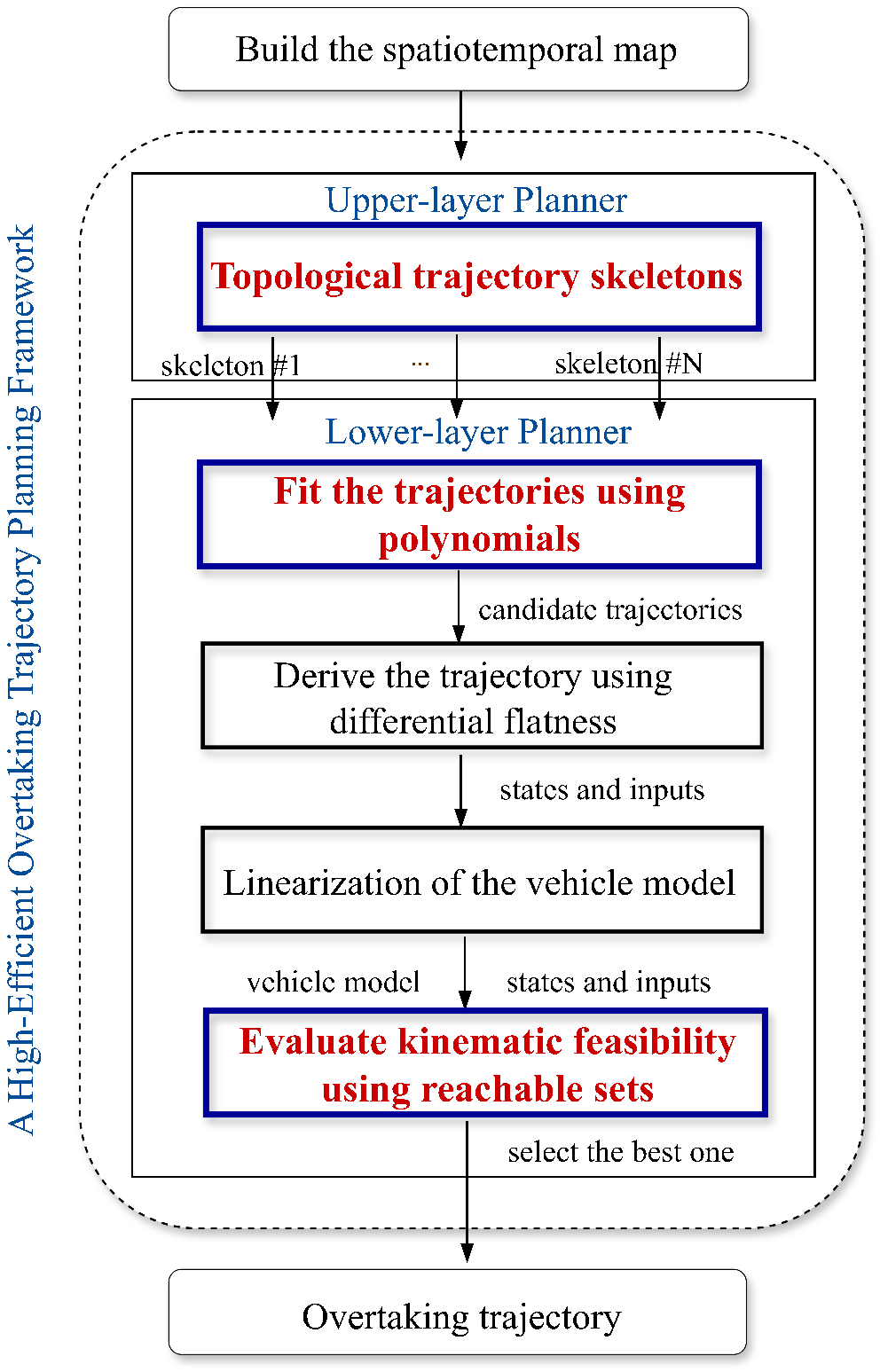}
  %% Use \caption command for figure caption and label.
  \caption{Overall architecture of the overtaking trajectory planning framework}\label{pipeline_fig}
  %% https://en.wikibooks.org/wiki/LaTeX/Importing_Graphics#Importing_external_graphics
\end{figure}

The upper layer planner in hierarchical planners generate a collision-free initial path, which serves as a reference for subsequent refinement. The main planning methods in upper layer planner include graph search-based methods, sampling-based methods, and their combinations. Specifically, search-based methods, such as A* \cite{hart1968formal} and Dijkstra \cite{sniedovich2006dijkstra}, are typically employed in partially known static environments. Improved algorithms based on the A*, D* algorithm \cite{stentz1994optimal} and Theta* algorithm \cite{daniel2010theta} can solve path planning problems in dynamic scenarios. In autonomous driving, A* searches a front-end collision-free trajectory in $s$-$l$-$t$ configuration space \cite{hesse2010motion}. Node expansion in \cite{ajanovic2018search} includes vertical and horizontal expansions. Different node combinations represent specific maneuvers, and the optimal trajectory is obtained through an A* search. Search-based methods find it difficult to balance the trade-off between trajectory quality and computation time precisely. As discretization precision increases, solution quality improves, but the computation time grows exponentially. Sampling-based methods are also common approaches for generating initial paths~\cite{kavraki1996probabilistic,mao2025rapid,zhao2026vision}. They usually involve random sampling in the state space. Classical random sampling methods include the Probabilistic Roadmap (PRM) algorithm  \cite{kavraki1996probabilistic} and the Rapidly-exploring Random Tree (RRT) algorithm \cite{lavalle2001randomized}. These methods randomly sample new nodes in the state space, connecting the new sample points with candidate neighbor nodes to form edges for subsequent searches. The trajectory edges between sample nodes can be generated by curve fitting or by solving optimization problems \cite{ziegler2009spatiotemporal, mcnaughton2011motion, werling2012optimal}.

\revmwl{However, both search-based and sampling-based planning methods typically provide a single initial solution, which may cause subsequent trajectory refinement to become trapped in a local optimum.} To overcome this limitation, some research has introduced the concept of path homotopy classes. Obstacles divide the entire planning space into different path topology classes. Each distinct topological trajectory represents an independent decision maneuver~ \cite{rosmann2017integrated}. \cite{yi2018model} proposed a method to divide the trajectory space into homotopy regions. An efficient topological path searching algorithm is proposed in \cite{zhou2020robust}, discovering multiple topological distinct paths in three-dimensional (3D) environments. \cite{rosmann2017integrated} maintains and optimizes acceptable candidate trajectories of different topologies according to the obstacle environment to seek the best solution. In a 3D static environment, \cite{ye2020tgk} finds a kinematically feasible trajectory through optimization methods from the start to the goal, then translates it in both directions until it does not collide with obstacles. These topological paths serve as reference paths for subsequent kinodynamic RRT* growth.  \cite{de2024topology} considers time information when searching for different homotopies to handle dynamic obstacles, which distinguishes it from \cite{zhou2020robust} and \cite{ye2020tgk}.

The lower layer planner in hierarchical planners generates a feasible trajectory for the vehicle based on the initial path. Optimization-based methods in the lower layer planner are commonly used for trajectory refinement. In \cite{ziegler2014trajectory}, static and dynamic obstacles are segmented using polygons, and then sequential quadratic programming methods are used to generate the trajectory. Gu et al. \cite{gu2013focused} sample an initial path in the state space and then optimize the selected curve further to obtain a feasible trajectory. \cite{qian2016optimal} uses a Mixed-Integer Quadratic Program to formulate trajectory planning while considering multiple logical constraints on the road. The literature ~\cite{han2023efficient,zhao2026vision} utilizes signed distance approximations to create constraints on other obstacles. Model predictive control (MPC) methods are also commonly used for trajectory refinement. Some linearized Model Predictive Control (LMPC) methods can simplify trajectory planning problems. For instance, \cite{liniger2015optimization,jain2019reacting,lattarulo2018linear} linearize the vehicle model and constraints to speed up solving the problem. However, this linearization can cause significant model errors, which may result in collisions. In contrast, nonlinear MPC (NMPC) methods can handle more complex optimization problems. \cite{subosits2019racetrack} integrates a dynamic vehicle model, while \cite{gritschneder2018fast} considers the constraint of trajectory curvature. These nonlinear terms increase the complexity of the problem. Nevertheless, these NMPC models can lead to longer solution times, making them unsuitable for scenarios that require high real-time performance. \revmwl{Model Predictive Contouring Control (MPCC) is also an effective optimal control method that simplifies the optimization problem by approximating the projection point onto the reference trajectory. In ~\cite{li2024reduce,li2025data}, velocity optimization was further incorporated to improve task performance. However, the effectiveness of these approaches in handling multi-agent interaction scenarios, such as overtaking, has not been thoroughly validated.} Optimization-based methods struggle to balance computational efficiency with solution accuracy. Simplified optimization problems improve computational efficiency but may produce infeasible trajectories. In contrast, complex optimization problems can ensure trajectory kinematic feasibility, yet they are highly computationally inefficient.

By temporarily disregarding the vehicle model in trajectory generation, time efficiency can be significantly improved. This can be accomplished by utilizing the reachable set of vehicles. The reachable set refers to the collection of all possible states a dynamic system can achieve under bounded constraints, starting from an initial state. In \cite{althoff2014online}, the reachable set is the collection of motion states that a vehicle can reach at any given moment while satisfying kinematic constraints. The reachable set provides a continuous drivable area. Areas outside this set represent states that do not meet the kinematic limits of vehicles. To enhance computational efficiency,  \cite{sontges2017computing} simplify the vehicle model to a point mass model to compute the reachable sets. The drivable corridor at any moment is formed by the union of the Cartesian products of the $x$ and $y$ directions, resulting in a 2D convex polytope. \cite{wursching2021sampling} effectively reduces the search space and significantly lowers the dimensionality of sampling by the reachable set. \cite{manzinger2020using,liu2024spatiotemporal} extract the driving corridors and generate a trajectory within these polygons using convex optimization methods. \cite{jewell2021embedded} integrate the kinematic vehicle model and account for uncertainties in state and input, forming a series of reachable sets. Since the reachable set represents all states that the vehicle can achieve under constrained inputs, we can evaluate the kinematic feasibility of a trajectory by examining the relationship between the trajectory and the corresponding reachable set.

\begin{figure}[!t]
  \begin{minipage}[t]{0.45\textwidth}
    \centering
    \includegraphics[width=\textwidth]{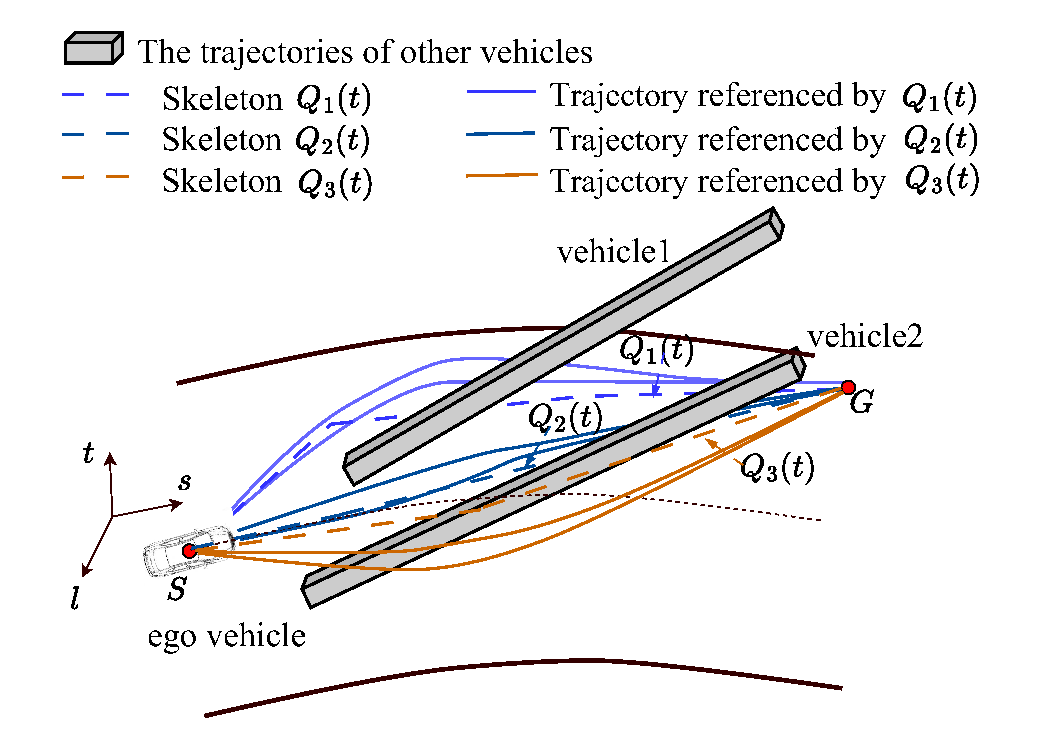}
    \subcaption{Illustration of overtaking trajectory generation in the
      $s$-$l$-$t$ configuration space}
    \label{overtake1}
  \end{minipage}
  \begin{minipage}[t]{0.45\textwidth}
    \centering
    \includegraphics[width=\textwidth]{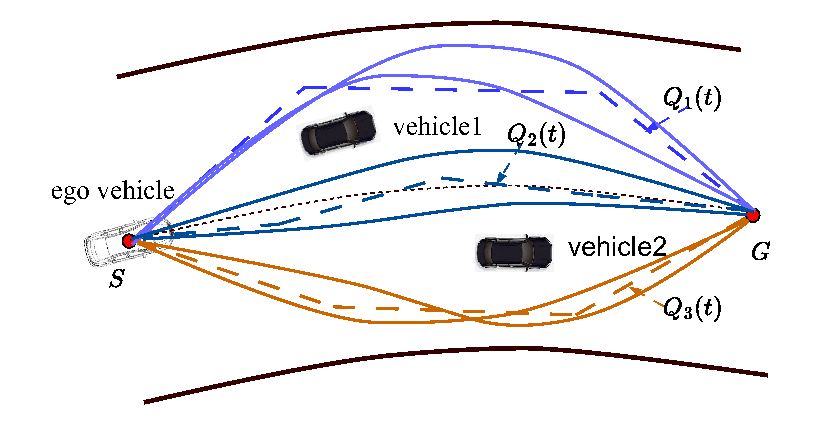}
    \subcaption{Illustration of overtaking trajectory generation on the plane}
    \label{overtake2}
  \end{minipage}
  \caption{Illustration of overtaking trajectory generation on the road. (a) Dashed lines represent trajectory skeletons from the upper-layer planner. Blue, purple and brown dashed lines belong to different topological classes. The solid lines show polynomial trajectories generated by the lower-layer planner, fitting based on the corresponding trajectory skeletons. Trajectories of the same color belong to the same topological class. (b) Distinctive topological trajectory represents a unique overtaking maneuver.}
  \label{fig:overtake}
\end{figure}

This paper proposes an overtaking trajectory planning framework based on spatiotemporal topology and reachable set analysis, to improve the trajectory quality and time efficiency. Specifically, the upper-layer planner searches for reference path skeletons of different topological types. These skeletons provide diverse initial solutions for refinement of the lower-layer planner, helping to avoid local optima and improve trajectory quality. The lower-layer planner introduces the reachable set to assess trajectory kinematic feasibility in parallel. Essentially, this method decouples trajectory optimization from the kinematic model through the reachable set, significantly reducing the complexity of the trajectory optimization problem and improving the efficiency of the solution. Figure \ref{fig:overtake} displays a schematic diagram illustrating trajectory planning for different topological classes in a scenario involving two other vehicles on the road.

\section{Preliminaries}
\label{preliminaries}
The reachable set of a vehicle represents the set of all continuous state trajectories that the vehicle can reach under the influence of the control variable within time. For a system $\dot \xi=f(\xi,u)$, the method for calculating the reachable set is as follows \cite{althoff2014online}:

\begin{table}[!t]
  \centering
  \caption{Notations in the spatiotemporal topological search method}
  \label{notation_table}
  \footnotesize
  \setlength{\tabcolsep}{14pt}
  \begin{tabular}{ll}
    \toprule %[2pt]   
    \textbf{Symbol} & \textbf{Description}                      \\
    \midrule %[2pt]
    $layer_i$       & the $i$th layer                           \\
    $\Delta s$      & the longitudinal sampling interval        \\
    $\Delta l$      & the lateral sampling interval             \\
    $\Delta t$      & the time sampling interval                 \\
    $p.e$           & the path skeleton from $p_s$ to $p$       \\
    $p_s$           & the starting node                         \\
    $p_g$           & the ending nodes                          \\
    $(p.s,p.l)$     & the coordinates of $p$ in Frenet frame    \\
    $(p.x,p.y)$     & the coordinates of $p$ in Cartesian frame \\
    $p.t$           & The time to reach node $p$ from $p_s$     \\
    $p.E_1$         & the edges set ending at $p$               \\
    $p.E_2$         & the edges set starting form $p$           \\
    $p.\mathcal{E}$ & all path skeletons from the $p_s$ to $p$  \\
    $T_{Ii}$        & the sampling time interval of $layer_i$   \\
    $t_i$           & the sampling start time of $layer_i$      \\
    $\mathcal{C}_i$ & the $i$th $link$ node set                 \\
    $v_m$           & the maximum velocity                      \\
    \bottomrule %[2pt]     
  \end{tabular}
\end{table}
The initial state is denoted by $\mathcal{R}(0)$, the bounded input is $\mathcal{U}$. For the initial state $\xi(0)=\xi_0,t\in[0,t_f]$, and trajectory $\mathcal{T}(\cdot)$, the state in reachable set is denoted by $\mathcal{X}(t,x_0,\mathcal{T}(\cdot),u(\cdot))$. The exact reachable set for a reference trajectory $\mathcal{T}^\star(\cdot)$ is
\begin{equation}
  \begin{split}
    \mathcal R^e([0,t_f]) = &\bigg\{\mathcal{X}(t,\xi_0,\mathcal{T}(\cdot),u(\cdot)) \big| t\in [0,t_f],\\
    &\xi_0\in\mathcal{R}(0),\mathcal{T}(t)=\mathcal{T}^\star(t),u(t)\in \mathcal{U}\bigg\}
  \end{split}
\end{equation}

Combine the state and input vector to a new vector $z=[\xi,u]$. For simplicity, using a first-order Taylor expansion around the linearization point $z^\star=[\xi^\star,u^\star]$, for $\forall  t \in [\tau_k,\tau_{k+1}]$
\begin{equation}
  \dot \xi_i \in f(z^\star)+ \frac{\partial f}{\partial z}\bigg|_{z=z^\star}\oplus\mathcal{L}_i(\tau_k)
\end{equation}
\begin{equation}
  \frac{\partial f}{\partial z}\bigg|_{z=z^\star}(z-z^\star)=\frac{\partial f}{\partial \xi}\bigg|_{\xi=\xi^\star}(\xi-\xi^\star) + \frac{\partial f}{\partial u}\bigg|_{u=u^\star}(u-u^\star)
\end{equation}
where $\oplus$ represents set-based addition \cite{althoff2014online}, $\mathcal{L}$ is the set of Lagrange remainders:
\begin{equation}
  \begin{split}
    \mathcal{L}_i(\tau_k)=&\bigg\{\frac{1}{2}(z-z^\star)^T\frac{\partial^2 f}{\partial z^2}\bigg|_{z=z^\star}(z-z^\star)\\
    &\bigg|(\eta=\mathcal{R}(\tau_k)\times \mathcal{U},z=\mathcal{R}(\tau_k)\times \mathcal{U})\bigg\}
  \end{split}
\end{equation}

Starting from $\mathcal{R}(t_k)$, compute the set of all solutions $\mathcal{R}_h(t_k+1)$ for the linearized system $\dot \xi = A\xi + Bu$. Using $r = t_{k+1}-t_k$, the solution of $\mathcal{R}_h(t_{k+1})$ is as follows:
\begin{equation}
  \begin{split}
    \mathcal{R}_h^d(t_{k+1})&=e^{Ar}\mathcal{R}(t_k)+x_p(r)\\
    &=e^{Ar}\mathcal{R}(t_k)+\int_{0}^{\tau}e^{A(r-t)}dt u_c
  \end{split}
\end{equation}

An interval matrix $\epsilon_p(r)=[-W(r)r,W(r)r]$ is introduced in \cite{althoff2011reachable}, then the particular solution $x_p(r)$ is bounded by
\begin{equation}
  \begin{split}
    x_p(r)\in \Gamma(r)\otimes u_c
  \end{split}
\end{equation}
where $\otimes$ represents set-based multiplication \cite{althoff2014online},  $\Gamma(r)$ is as follows:
\begin{equation}
  \begin{split}
    \Gamma(r)=\sum_{i=0}^{\gamma}\frac{A^ir^{i+1}}{(i+1)!}\oplus\epsilon_p(r)
  \end{split}
\end{equation}

Considering the uncertainty of input $U_r$, the reachable set is
\begin{equation}
  \begin{split}
    \mathcal{R}_p(r)=\bigg[\sum_{i=0}^{\gamma}\frac{A^ir^{i+1}}{(i+1)!}\otimes U_r\bigg]\oplus\bigg[\epsilon_p(r)\otimes \left\vert U_r \right\vert\bigg]
  \end{split}
\end{equation}

The enlargement required to bound all affine solutions within $\tau_k$ is denoted by $\mathcal{R}_\epsilon$
The reachable set for the next point in time and time interval is obtained by combining all previous results and using the operator $co(\cdot)$ for the convex hull:
\begin{subnumcases} {\label{reachable_set}}
  \mathcal{R}(t_{k+1})=e^{Ar}\mathcal{R}(t_k)\oplus \Gamma(r)u_c\label{reach_set1}\\
  \mathcal{R}(\tau_k)=co(\mathcal{R}(t_k),\mathcal{R}(t_{k+1}))\oplus \mathcal{R}_\epsilon \oplus \mathcal{R}_p(r)\label{reach_set2}
\end{subnumcases}

Based on the above method, the reachable set of the vehicle system at any given time can be calculated given the initial state and control inputs.
\section{The Overtaking Trajectory Planning Framework}
\label{sec_STHPTP}
In this chapter, the implementation details of the overtaking trajectory planning framework based on spatio-temporal topology and reachable set analysis are introduced. The overall structure of the algorithm is shown in Figure \ref{pipeline_fig}.
\subsection{Overall Framework}
\label{sec_overallframework}
The overtaking trajectory planning framework includes the upper-layer planner based on \textbf{s}patiotemporal \textbf{t}opological \textbf{s}earch method (STS)\label{STS}. And the lower-layer planner based on \textbf{p}arallel \textbf{t}rajectory \textbf{g}eneration using \textbf{r}eachable sets (RPTG)\label{RPTG}. The upper-layer planner translates the trajectories of other vehicles on the road into the $s$-$l$-$t$ configuration space, treating them as static obstacles. It performs sampling in the $s$-$l$-$t$ configuration space and applies graph search methods to identify a set of collision-free initial trajectories from different topological classes. The lower-layer planner concurrently optimizes these initial trajectories to derive time-parameterized polynomial candidate trajectories. Leveraging the differential flatness property of the vehicle, the inputs of can be determined at specific time instances. The reachable sets method is then used to assess the kinematic feasibility of the candidate trajectories. The optimal collision-free and highly control-feasible trajectory is selected as the overtaking trajectory, which is provided to the controller for tracking.

\subsection{Spatiotemporal Topological Search Method}
\label{sec_upperplanner}

In the spatiotemporal configuration space, the trajectory of other moving vehicles is regarded as a static obstacle, dividing the entire search space into independent topologies. This section aims to search within these topological spaces to obtain a series of path skeletons that represent different motion behaviors, providing references for subsequent trajectory generation.
\subsubsection{Notation Description}
\begin{figure}[!t]
  \centering
  \begin{minipage}[c]{0.45\textwidth}
    \centering
    \includegraphics[width=\textwidth]{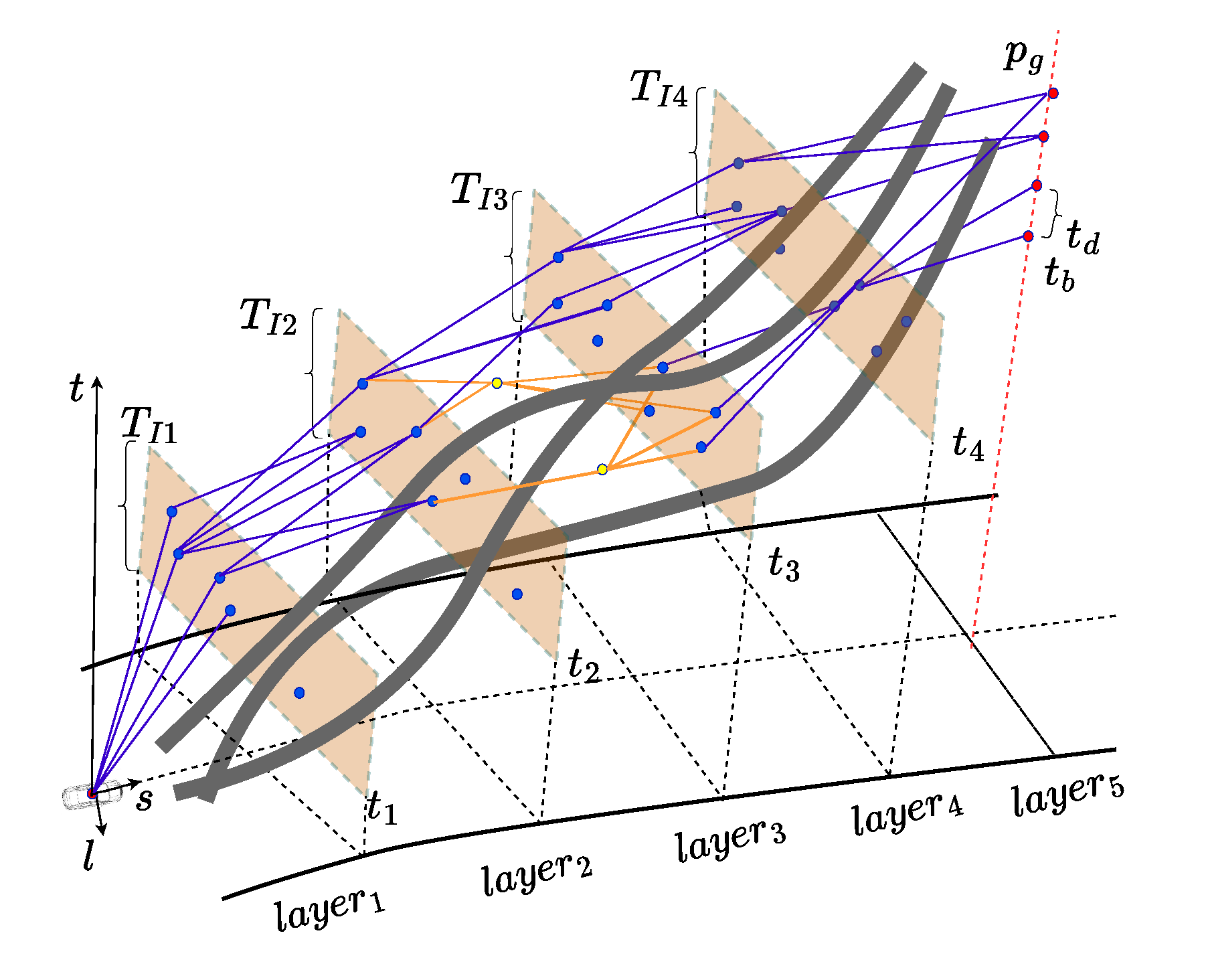}
    \subcaption{Road search graph in spatio-temporal space. }
    \label{fig_sample_1}
  \end{minipage} \\
  \begin{minipage}[c]{0.45\textwidth}
    \centering
    \includegraphics[width=\textwidth]{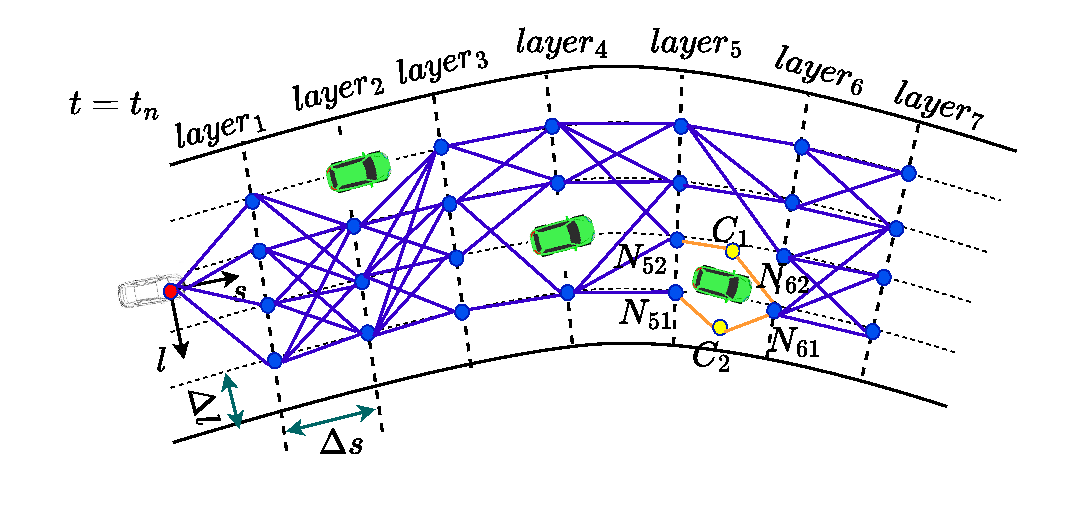}
    \subcaption{The search space on raod at $t=t_n$. }
    \label{fig_sample_2}
  \end{minipage}
  \caption{Road skeleton search graph. In (a), the red nodes represent the start and end nodes. The blue nodes are layer nodes, and the yellow nodes are link nodes. The sampling time interval for $layer_i$ is $[t_i,t_i+T_{Ii}]$. The brown boxes indicate the sampling range for each $layer_i$. The blue edges represent direct connections between layer nodes from adjacent layers. In (b), due to obstacles blocking the connections, it forms new edges:$\{p_{51},p_{l1},p_{61}\},\{p_{52},p_{l2},p_{62}\}$.}
  \label{fig_sample}
\end{figure}

As shown in Figure \ref{fig_sample_1}, the spatiotemporal sampling space consists of the plane and time, the sampling node $p = \mathbb{R}^2 \times [0,T]$. The trajectory of $j$th vehicles is regarded as an obstacle, $j\in \mathbb{N}_0$, the occupancy states set is represented by the thick black lines. In this section, the Frenet coordinate system \cite{werling2012optimal} is constructed along the road centerline. Table \ref{notation_table} explains some symbols in this section.

The search graph shown in Figure \ref{fig_sample_1} contains two types of sampling nodes: the $layer$ $node$ and $link$ $node$. The position of the $layer$ $node$ is obtained by sampling on the corresponding layer. Several link nodes are sampled in the free area between adjacent layers. \revmwl{When obstacles block the connections between the adjacent layers nodes, they can be bypassed using link nodes (highlighted as yellow edges in Figure \ref{fig_sample_2})}. It is worth noting that $p_g$ contains a series of nodes with the exact geographic coordinates but differ in arrival times. Starting at node $p$, they backtrack through $p.E_2$ and $p.E_1$, recording each layer node and link node traversed until reaching $p_s$. This process constructs a path skeleton $e$ from $p_s$ to $p$.

\subsection{Search Referece Topological Skeletons}
To extract candidate reference skeletons, traverse from the first layer, sampling a series of link nodes in the free space between every two adjacent layers. Traverse all nodes in the layer, attempting to connect with nodes in the next layer. Nodes within the same layer are not allowed to connect. If the line between two layer nodes is unobstructed and meets the maximum speed limit, these nodes can form an edge. If an obstacle obstructs the line between two nodes, find a suitable link node in the corresponding set $\mathcal{C}$ and connect it to aforementioned two-layer nodes. These three nodes form an edge if they are not obstructed and meet the maximum speed limit. Add the newly generated edge to the edges set of the corresponding layer nodes, then update the cost of each skeleton. Repeat the above process until all the edge sets of $p_g$ are updated. Add all skeletons from $p_s$ to $p_g$ to the priority queue according to cost in ascending order. Finally, skeletons with the lowest costs, each belonging to a different topological class, are extracted from the priority queue. The general structure is outlined in Algorithm \ref{alg1}.

\begin{algorithm}[t]
	\renewcommand{\algorithmicrequire}{\textbf{Input:}}
	\renewcommand{\algorithmicensure}{\textbf{Output:}}
	\caption{Spatiotemporal Graph Construction}
	\label{alg1}
	\begin{algorithmic}[1]
		\STATE \algorithmicrequire\ $\mathbf{E}$: road map, $N_s$: maximum sample number, $N_p$: number of skeletons
		\STATE \algorithmicensure\ $\mathcal{E}_r$: set of skeletons
		\STATE $\mathcal{E}_r \leftarrow \emptyset, N_s \leftarrow 0, N \leftarrow (p_g.s - p_s.s)/\Delta s, i \leftarrow 0$
		\FOR {\textbf{each} $layer_i$}
		\STATE sample $N_s$ nodes and add them to $\mathcal{C}_i$
		\FOR {\textbf{each} $p_1 \in layer_i$}
		\FOR {\textbf{each} $p_2 \in layer_{i+1}$}
		\IF {$\lnot$ \textbf{LineVisible}($p_1,p_2$)}
		\STATE find the first feasible link node $p_l$ in $\mathcal{C}_i$
		\STATE $p_1.E_2, p_2.E_1 \overset{+}\leftarrow \{p_1,p_l,p_2\}$
		\ELSE
		\IF {\textbf{IsReachable}($p_1,p_2$)}
		\STATE $p_1.E_2, p_2.E_1 \overset{+}\leftarrow \{p_1,p_2\}$
		\ENDIF
		\ENDIF
		\STATE \textbf{UpdateSkeletonsCost}($p_1$)
		\ENDFOR
		\ENDFOR
		\ENDFOR
		\STATE Add all elements of $p_g.E$ to the priority queue $pq$
		\STATE $\mathcal{E}_r \leftarrow$ \textbf{ExtractDistinctSkeletons}($pq,N_p$)
		\RETURN $\mathcal{E}_r$
	\end{algorithmic}
\end{algorithm}

The distance between $p_1$ and $p_2$ is estimated by (\ref{dist}), denoted by $dist(p_1,p_2)$.
\begin{equation}
  dist(p_1.p_2) =  \sqrt{(p_2.s-p_1.s)^2+(p_2.l-p_1.l)^2}\label{dist}
\end{equation}

\textbf{IsReachable}($p_1,p_2$) is used to determine whether $(p_1,p_2)$ is reachable. $(p_1,p_2)$ is reachable when $p_2.t-p_1.t\ge dist(p_1,p_2)/v_m$. This indicates that $p_2$ from $p_1$ can be reached without exceeding the maximum velocity $v_m$.

If the line between $p_1$ and $p_2$ is blocked by an obstacle, traverse the elements in $\mathcal{C}_i$ until a link node $p_l$ is found such that the line between $p_1$ and
$p_l$, and the line between $p_l$ and $p_2$ both are unobstructed and are reachable.

\begin{figure}[!t]
	\begin{minipage}[t]{0.22\textwidth}
		\centering
		\includegraphics[width=1.45in]{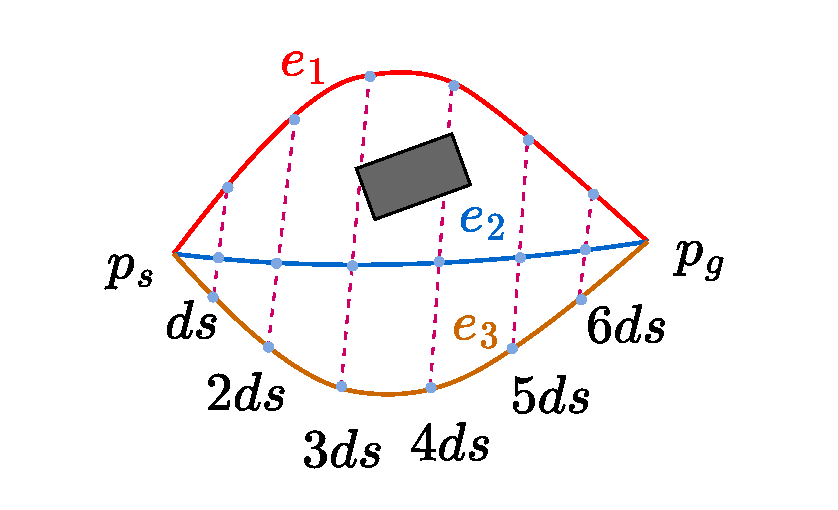}
		\subcaption{}
		\label{UVD1}
	\end{minipage}
	\begin{minipage}[t]{0.22\textwidth}
		\centering
		\includegraphics[width=1.45in]{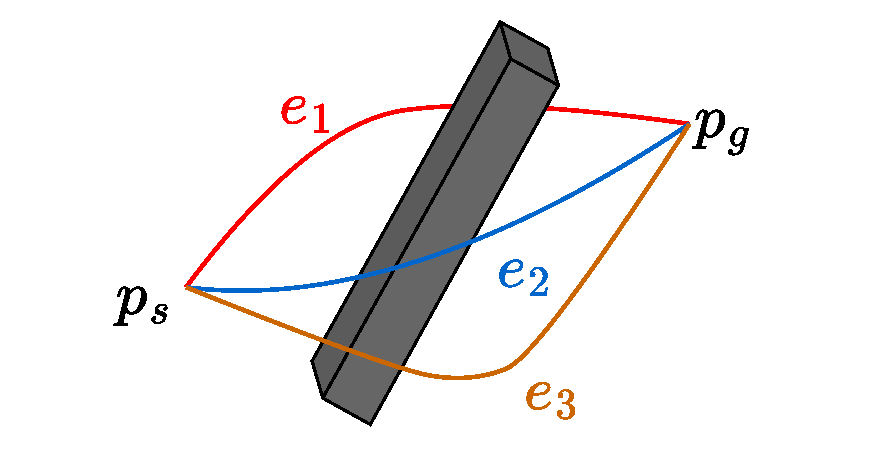}
		\subcaption{
		}
		\label{UVD2}
	\end{minipage}
	\caption{Skeletons topology equivalence. (a) Starting from $p_s$, a point is sampled every $ds$. If the line connecting the corresponding points is obstructed, it indicates that the two skeletons are topologically non-equivalent. (b) An illustration of topology equivalence, $e_2$ and $e_3$ are topological equivalents while the topology of $e_1$ is distinct to $e_2$ and $e_3$.}
	\label{fig:UVD}
\end{figure}

The calculation of the cost of path skeletons will be described in detail later. After adding all the skeletons from $p_s$ and $p_g$ to the priority queue $pq$, only a small number $N_p$ of skeletons with the smallest costs and different topologies are selected from $pq$ as reference path skeletons. Uniform visibility deformation(UVD) proposed in \cite{zhou2020robust} is used to determine the topological equivalence of the two skeletons in this section, as shown in Figure \ref{fig:UVD}.

\subsubsection{Cost Fucntion}
The cost of the path skeleton is calculated in the function \textbf{UpdateSkeletonsCost}, as shown in equation (\ref{edgecost}).
\begin{equation}
  J_{\mathcal{E}} =  c_1J_{T} +c_2J_\theta +c_3J_{len}+c_4J_{acc}+c_5J_{obs}\label{edgecost}
\end{equation}
where $J_T$ represents the time cost of reaching the endpoint along the path skeleton, $J_\theta$ represents the curvature cost of the path skeleton, $J_{acc}$ represents the acceleration cost of the path skeleton, $J_{len}$ represents the length cost of the path skeleton, and $J_{obs}$ represents the proximity cost of the path to obstacles. For the path skeleton $\mathcal{E}$, which contains $N+1$ nodes, $p_0$ and $p_{N+1}$ are the starting and ending nodes of the skeleton, respectively. The specific calculation method for evaluating the cost of skeleton $\mathcal{E}$ is as follows:
\begin{numcases}{}
  t_{ref} = \frac{dist(p_0, p_{N+1})}{v_m} \label{refT} \\
  J_T \quad = \frac{p_e t}{t_{ref}} \label{cost_time}
\end{numcases}
\begin{equation}
  J_{len} = \sum_{i=1}^N\frac{\lVert \overrightarrow{p_{i-1}p_i}\rVert}{dist(p_0,p_{N+1})}\label{cost_len}
\end{equation}
(\ref{cost_time}) and (\ref{cost_len}) calculate the time cost and length cost of the path skeleton, respectively.
(\ref{calc_bending}) calculates angles between any three consecutive nodes $p_{i-1},p_i,p_{i+1}$ by computing the angle formed between the vectors $\overrightarrow{p_{i-1}p_i}$ and $\overrightarrow{p_{i}p_{i+1}}$.
\begin{numcases}{}
  \theta_i = \arccos(\frac{\overrightarrow{p_{i-1}p_i} \cdot \overrightarrow{p_{i}p_{i+1}}}{\lVert \overrightarrow{p_{i-1}p_i}\rVert \cdot \lVert \overrightarrow{p_{i}p_{i+1}}\rVert})\label{calc_bending}\\
  J_{\theta}=\sum_{i=1}^N\frac{\theta_i}{\pi}\label{cost_theta}
\end{numcases}

This paper explores using standard deviation analysis on a series of acceleration measurements to assess the stability of acceleration variation. A smaller standard deviation indicates smoother changes in acceleration, which may correlate with higher comfort levels. $\Delta t_{i-1}$ represents the time deviation between $p_i$ and $p_{i-1}$ while $\Delta t_{i}$ represents the time deviation between $p_{i+1}$ and $p_i$. The average acceleration across all points is denoted as $\bar{a}$, $a_i$ is the acceleration at node $p_i$.
\begin{numcases}{}
  a_i = \frac{2(\lVert \overrightarrow{p_ip_{i+1}}\rVert/\Delta t_i-\lVert \overrightarrow{p_{i-1}p_i}\rVert/\Delta t_{i-1})}{(\Delta t_{i-1}+\Delta t_i)}\label{acc}\\
  J_{acc} = \sqrt{\frac{1}{N}\sum_{i=1}^N(a_i-\bar{a})^2}\label{cost_acc}
\end{numcases}

As defined in Equation (\ref{cost_obs}), a penalty function is used to penalize points on the trajectory that come too close to obstacles. Points located within a threshold distance $r_{th}$ from obstacles are subject to a penalty that increases as the distance decreases. This penalty function encourages trajectories to maintain a safe distance from obstacles. A total of $M$ points are sampled uniformly along the skeleton.
\begin{equation}
  J_{obs} = \sum_{i=1}^M\frac{(r_{th}-\min(r_{th},r_i))}{M\cdot r_th}\label{cost_obs}
\end{equation}

In this section, a series of topological reference trajectories are generated in the $s$-$l$-$t$ configuration space. These topological reference trajectories, belonging to different topological classes, represent independent motion behaviors. Therefore, optimizing these reference skeletons as initial solutions can help avoid getting trapped in local optima, thereby improving the quality of the resulting trajectories. The lower-layer trajectory planner utilizes these topological reference skeletons as initial solutions and generates a set of collision-free candidate trajectories in parallel. The specific implementation details of the lower-layer trajectory planner will be discussed in the next section.

\subsection{Parallel Trajectory Generation Method Based on Reachable Set}
\label{sec_RSPT}
This section outlines a method for generating parallel feasible trajectories using reachable sets. The process begins by obtaining a series of reference skeletons from different topological classes. Then, the lower-layer trajectory planner optimizes each reference skeleton in parallel, ultimately selecting the optimal, control-feasible trajectory for overtaking. Initially, each reference skeleton is fitted using a quintic polynomial. The fitting problem is formulated as a quadratic programming problem, considering both the fitting cost of the quintic curve to the skeleton and the jerk cost of the curve. By adjusting the ratio between these two costs, a set of candidate fitting trajectories can be generated in parallel. Although these trajectories can be efficiently obtained through quadratic programming, this approach does not consider the kinematic model of the vehicle, and thus fails to satisfy kinematic constraints. To address this issue, the necessary inputs for the candidate trajectories are computed based on the differential flatness property. Finally, using these inputs, reachable sets are employed to identify the optimal, control-feasible trajectory from among the candidate curves, resulting in the final overtaking trajectory.
\subsubsection{Candidate Trajectories Generation}
\label{sec_CTG}
The trajectory skeletons generated in Section \ref{sec_upperplanner} will be fitted with polynomials to create a series of candidate trajectories in this section. These candidate trajectories will later be evaluated for collision-free status and kinematic feasibility.
The position of the vehicle can be described by a polynomial with degree $n$, $Q(t) = \mathbf{k}^T\beta(t)$, $t\in [0, T]$ where the coefficient vector $\mathbf{k}\in\mathbb{R}^{2n}$, $T$ is the time duration, $\beta(t)=(1,t,t^2,\cdots,t^{2n-1})^T$ is the natural basis. $Q^\star(t)$ is the path skeleton. The above trajectory fitting problem can be formulated as an optimization problem in the following form:
\begin{equation}
  \begin{split}
    &\min \,\, J\\
    &s.t.\quad  \begin{array}{lc}
      Q^{(j)}(T)=Q^{\star(j)}(T) \\
      Q^{(j)}(0)=Q^{\star(j)}(0),j\in\{0,1,2\}
    \end{array}
  \end{split}\label{candidate_opti}
\end{equation}
where the constraints include position, velocity and acceleration constraints at the start and goal. The form of the objective function in (\ref{candidate_opti}) is as follows.
\begin{equation}
  J={\alpha_1\int_{0}^{T}\left\|Q(t)-Q^\star(t)\right\|^2dt}+{\alpha_2\int_{0}^{T}\left\|\dddot{Q}(t)\right\|^2dt}\label{refine_traj}
\end{equation}

(\ref{refine_traj}) consists of two parts: the first part is the cost of deviation between the polynomial trajectory and the path skeleton, and the second part is the cost of the smoothness of the polynomial trajectory.
Since the optimization problem only includes equality constraints, it can be rewritten as an unconstrained quadratic optimization problem, which can be solved fast \cite{richter2016polynomial}. $\alpha_1$ and $\alpha_2$ are the weights.

Since the original trajectory skeleton $Q^\star$ is collision-free, increasing $\alpha_1$ makes the polynomial trajectory closer to $Q^\star$. However, this sacrifices smoothness and might make the trajectory infeasible. \revmwl{Conversely, while increasing $\alpha_2$ yields a smoother trajectory, it exacerbates the spatial deviation from $Q^\star$, thereby escalating collision risks.} To determine the best combination of $\alpha_1$ and $\alpha_2$, this paper explores various pairs of $\alpha_1$ and $\alpha_2$, generating multiple candidate trajectories for each skeleton. These trajectories are then evaluated for collision status and kinematic feasibility to determine the optimal trajectory.
\subsubsection{Determine the Inputs Based on Differential Flatness}
To evaluate the kinematic feasibility of candidate trajectories, it is first necessary to determine the states and inputs of the vehicle along the trajectories. From the differential flatness property of the vehicle, it is known that the states and inputs can be derived from the position and their finite-order derivatives \cite{murray1995differential}. This section will illustrate deriving the states and controls from the polynomial trajectories.

For the kinematic bicycle model in Cartesian coordinates frame, the state vector $\mathbf{\xi} = (x,y,v,\theta)^T$, the input vector $\mathbf{u} = (a_t,\delta)^T$. Where $(x,y)^T$ is the position at the center of the rear wheels, $v$ is the longitudinal velocity, $a_t$ is the longitude acceleration, and $\delta$ is the steering angle. Based on the characteristics of differential flatness, all information is expressed through $(x,y)^T$, as shown in equation (\ref{df_eq}).
\begin{subnumcases} {\label{df_eq}}
  v = \sqrt{\dot{x}^2+\dot{y}^2}\label{df_v}\\
  \theta = \tan^{-1}(\dot{y}/\dot{x})\label{df_theta}\\
  a_t = (\dot{x}\ddot{x}+\dot{y}\ddot{y})/\sqrt{\dot{x}^2+\dot{y}^2}\label{df_a}\\
  \delta = \tan^{-1}((\dot{x}\ddot{y}-\dot{y}\ddot{x})L/(\dot{x}^2+\dot{y}^2)^{\frac{3}{2}})\label{df_delta}
\end{subnumcases}
where $L$ is the length of the wheelbase. Thus, the flat outputs $(x,y)^T$ and their finite derivatives describe the arbitrary state and input information of the vehicle.
\subsubsection{Evaluate Kinematic Feasibility Using Reachable Sets}
\label{sec_reachableset}
After obtaining the states and inputs of the ego car along the candidate trajectories, this section introduces the reachable sets to evaluate the kinematic feasibility of the candidate trajectories in parallel.
A continuous state space representation of the kinematic vehicle model can be expressed as:
\begin{subnumcases} {\label{bike_model}\dot \xi = f(\xi,u)\equiv}
  \dot x = \cos(\theta)\cdot v\label{model_x}\\
  \dot y = \sin(\theta)\cdot v\label{model_y}\\
  \dot \theta = \tan(\delta)/L\cdot v\\
  \dot v = a_t
\end{subnumcases}

According to the calculation of reachable sets described in section \ref{preliminaries}, the kinematic model in (\ref{bike_model}) is first linearized as follows:
\begin{equation}
  f(\xi,u)\cong f(
  \xi^\star,u^\star)+\frac{\partial f}{\partial \xi}\bigg|_{\xi=\xi^\star}(\xi-\xi^\star) + \frac{\partial f}{\partial u}\bigg|_{u=u^\star}(u-u^\star)\label{lin1}
\end{equation}
\begin{equation}
  \xi_{k+1} = \xi_k + f(\xi_k,u_k)\cdot dt\label{lin2}
\end{equation}
Convert the above formula into matrix form.
\begin{equation}
  \xi_{k+1} = A_k\xi_k + B_k u_k\label{lin3}
\end{equation}
\begin{equation}
  A_k =\begin{bmatrix} A_k^\star  & C_k^\star  \\
    \mathbf{0} & \mathbf{0}\end{bmatrix}
  \quad B_k =\begin{bmatrix}B_k^\star \\
    \mathbf{0}\end{bmatrix}\label{lin4}
\end{equation}
\begin{equation}
  A_k^\star = \frac{\partial f}{\partial \xi}\bigg|_{\xi=\xi^\star}\cdot dt\label{lin5}
\end{equation}
\begin{equation}
  B_k^\star = \frac{\partial f}{\partial u}\bigg|_{u=u^\star}\cdot dt\label{lin6}
\end{equation}
\begin{equation}
  C_k^\star = f(
  \xi^\star,u^\star)\cdot dt -A_k^\star \xi^\star -B_k^\star u_k^\star\label{lin7}
\end{equation}

where $\xi_k = \begin{bmatrix}\xi&\mathbf{I} \end{bmatrix}^T$. $\mathbf{0}$ and $\mathbf{I}$ are the zero matrix and identity matrix with appropriate dimensions.

The planning time horizon is the arriving time $T$ of the goal. The time increment is $\Delta T$, $T = N\Delta T$. A discrete step is denoted by $k \in \mathbb{N}_0$ , $k \in [0,N]$. The discrete-time at $k$, $t_k = k\Delta T$. Then the corresponding linearized model is obtained based on (\ref{lin2})-(\ref{lin4}) at each $t_k$. Starting from $t=0$, a linearized reference state $\xi^\star_k$ is selected on the candidate trajectory at intervals of $\Delta T$. From all states in $\mathcal{R}_k$, the system is simulated forward for a time interval $\Delta T$ using the linearized model at $\xi_k$, under the bounded inputs and the input uncertainty $U_r$. Thus the next reachable set $\mathcal{R}_{k+1}$ is then calculated by (\ref{reachable_set}). The initial reachable set $\mathcal{R}_0$ contains only one state $\xi_0$. This process repeats until the reachable set at the final time $T$ is determined. To describe a trajectory with high kinematic feasibility, the following definition is provided.

\begin{definition}[\textbf{High Kinematic Feasibility Trajectory}]Given the trajectory $\mathcal{T}(x,u)$ with the initial state $\xi_{0}^\star$ at time $t_0$. The linearization horizon is $N$, the linearization state of the trajectory is $\xi_{k}^\star$ at time $t_k$. The consecutive time intervals $\tau_k=[t_k,t_{k+1}]$, $\mathcal{T}(x,u)$ is regarded as kinematic feasible trajectory in $[t_0,t_{k+1}]$ if $\xi_k \in \mathcal{R}(\tau_k)$, $\forall k \in N$.
\end{definition}

\begin{figure}[!t]
  \begin{minipage}[t]{0.5\textwidth}
    \centering
    \includegraphics[width=\textwidth]{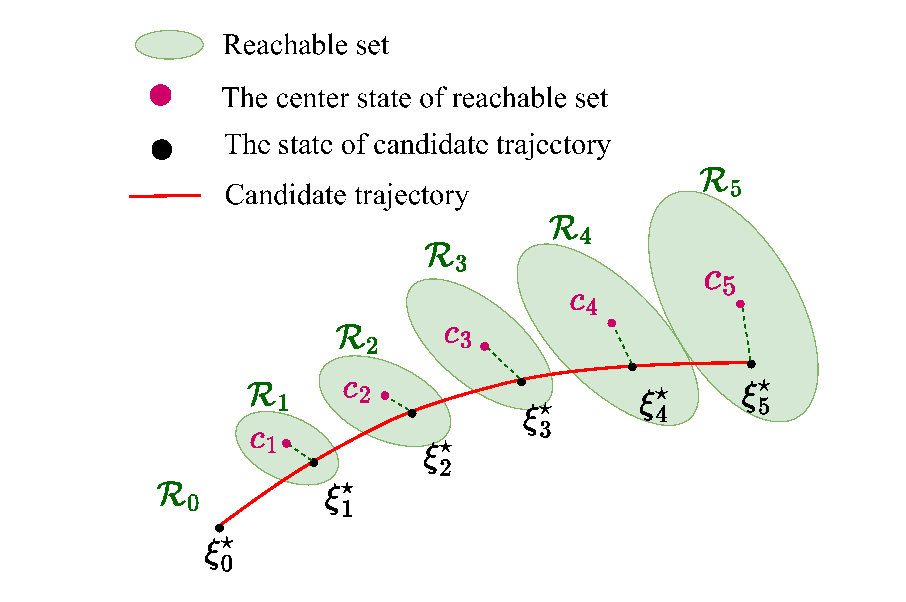}
    \subcaption{}
    \label{reachable1}
  \end{minipage}
  \begin{minipage}[t]{0.45\textwidth}
    \centering
    \includegraphics[width=\textwidth]{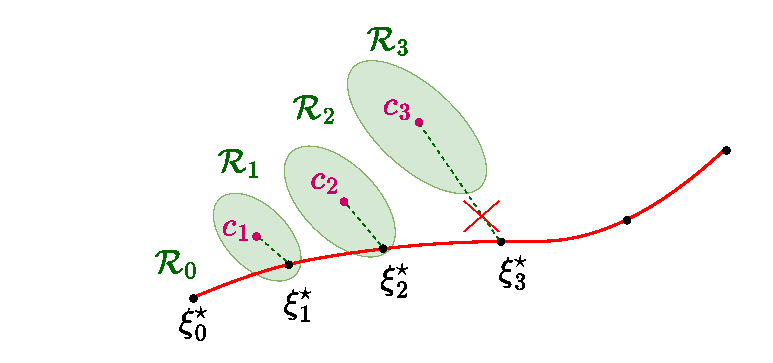}
    \subcaption{
    }
    \label{reachable2}
  \end{minipage}
  \caption{Illustration of using reachable sets to determine trajectory kinematic feasibility. (a) $\mathcal{R}_0$ is the reachable set of the vehicle at $t=0$, and it is a single state.  Along the trajectory, linearized reference states, such as $\xi^\star_0$-$\xi^\star_5$, are determined at intervals of $\Delta T$. In the diagram, each state is contained within its corresponding reachable set, indicating that the trajectory has high kinematic feasibility. (b) This diagram shows a low kinematic feasibility trajectory because $\xi^\star_3$ is no longer contained within its corresponding reachable set.}
  \label{fig:reachableset}
\end{figure}

Since it is impossible to enumerate the reachable sets at all times, the high kinematic feasibility of the candidate trajectories is approximately determined according to the method in Definition 1. Figure \ref{fig:reachableset} provides intuitive examples of trajectories with high kinematic feasibility and those with low kinematic feasibility, respectively.
There may be multiple trajectories with high kinematic feasibility, so it is necessary to select the optimal trajectory as the overtaking trajectory. Therefore, the kinematic feasibility of the trajectories is evaluated according to the following aspects:
\begin{align}
  J_{RS} & = J_p + J_v + J_{\theta} \nonumber                                                                                                              \\
         & = \frac{1}{N}\sum_{i=1}^N\left(\lambda_1\frac{p^d_k}{d_r} + \lambda_2\frac{v^d_k}{v_r} + \lambda_3\frac{\xi^d_k}{\theta_r}\right)\label{rscost}
\end{align}
where $J_p$\label{J_p} represents the average deviation cost of the trajectory points from the center of the corresponding reachable set in position; $J_v$\label{J_v} represents the average deviation cost of the trajectory states from the center of the corresponding reachable set in velocity; and $J_\theta$\label{J_theta} represents the average deviation cost of the trajectory states from the center of the corresponding reachable set in heading angle. $p^d_k$, $v^d_k$, and $\theta^d_k$ represent the position deviation, velocity deviation, and orientation angle deviation between the reference state $\xi^\star_k$ and the center $c_k$ of the corresponding reachable set, respectively. $\lambda_1$, $\lambda_2$, and $\lambda_3$ are coefficients, and $d_r$, $v_r$, and $\theta_r$ are reference values used to normalize the corresponding terms. Specifically,
\begin{equation}
  p^d_k = \sqrt{(c_k.x-\xi^\star_k.x)^2+(c_k.y-\xi^\star_k.y)^2}\label{rs_pd}
\end{equation}
\begin{equation}
  v^d_k = \lvert c_k.v-\xi^\star_k.v \rvert\label{rs_vd}
\end{equation}
\begin{equation}
  \theta^d_k = \lvert c_k.\theta-\xi^\star_k.\theta \rvert\label{rs_angled}
\end{equation}

The smaller the value of $J_{RS}$, the smaller the deviation between the reference state and the center of the reachable set, indicating stronger robustness and kinematic feasibility. The trajectory corresponding to the smallest $J_{RS}$ will be selected as the overtaking trajectory.

\section{Experimental Results and Discussion}
\label{Results}
In this section, we validate the proposed algorithm in simulation. To verify the performance and efficiency of the proposed method, we compared it with a common hierarchical planning approach, where the upper layer uses a dynamic programming (DP) approach \cite{bakker2005hierarchical}, and the lower layer employs a Nonlinear Model Predictive Control (NMPC) \cite{zuo2020mpc} trajectory optimization method. The algorithm was implemented in C++/Python, and the CPU used for the experiments is an AMD Ryzen 7 5800H running at 3.2 GHz.
\subsection{Simulation Setup}
\revmwl{The vehicle kinematic parameters for both the numerical experiments and the F1TENTH simulation are presented in Table \ref{tab:setting}. }The uncertainties in vehicle acceleration and front wheel angle arise from sensor uncertainties. To simplify the problem, we will assume these uncertainties to be constant values.
\begin{table}[htbp]
  \centering
  \caption{\revmwl{Simulation parameter settings}}
  \label{tab:setting}
  \footnotesize
  \setlength{\tabcolsep}{2pt}
    \begin{tabular}{llcc}
      \toprule
      
      \multirow{2}{*}{Parameter} & \multirow{2}{*}{Description} & \multicolumn{2}{c}{Value} \\
      \cmidrule(lr){3-4}
      & & Numerical & F1TENTH \\
      \midrule
      
      $v_{m}$ (m/s)        & Maximum velocity     & 15               & 3 \\
      $a_m$ (m/s$^2$)      & Maximum acceleration & 5                & 5 \\
      Length (m)           & The length of car    & 4.3              & 0.45 \\
      Width (m)            & The width of car     & 1.9              & 0.2 \\
      $L$ (m)              & The wheelbase        & 2.8              & 0.32 \\
      $\delta_m$ (rad)     & Maximum steer angle  & 0.52          & 0.42 \\
      $r_{th}$ (m)         & The safe distance    & 0.1              & 0.05 \\
      $U_r$ (m/s$^2$, rad) & Input uncertainties  & $[0.01, 0.005]$  & $[0.05, 0.01]$ \\
      
      \bottomrule
    \end{tabular}%

\end{table}

\subsection{On the Effectiveness and Efficiency of Our Method}

\subsubsection{The effect of different fitting parameters on trajectory quality}
\begin{figure}[!t]%% placement specifier
  \centering%% For centre alignment of image.
  \includegraphics[width=0.45\textwidth]{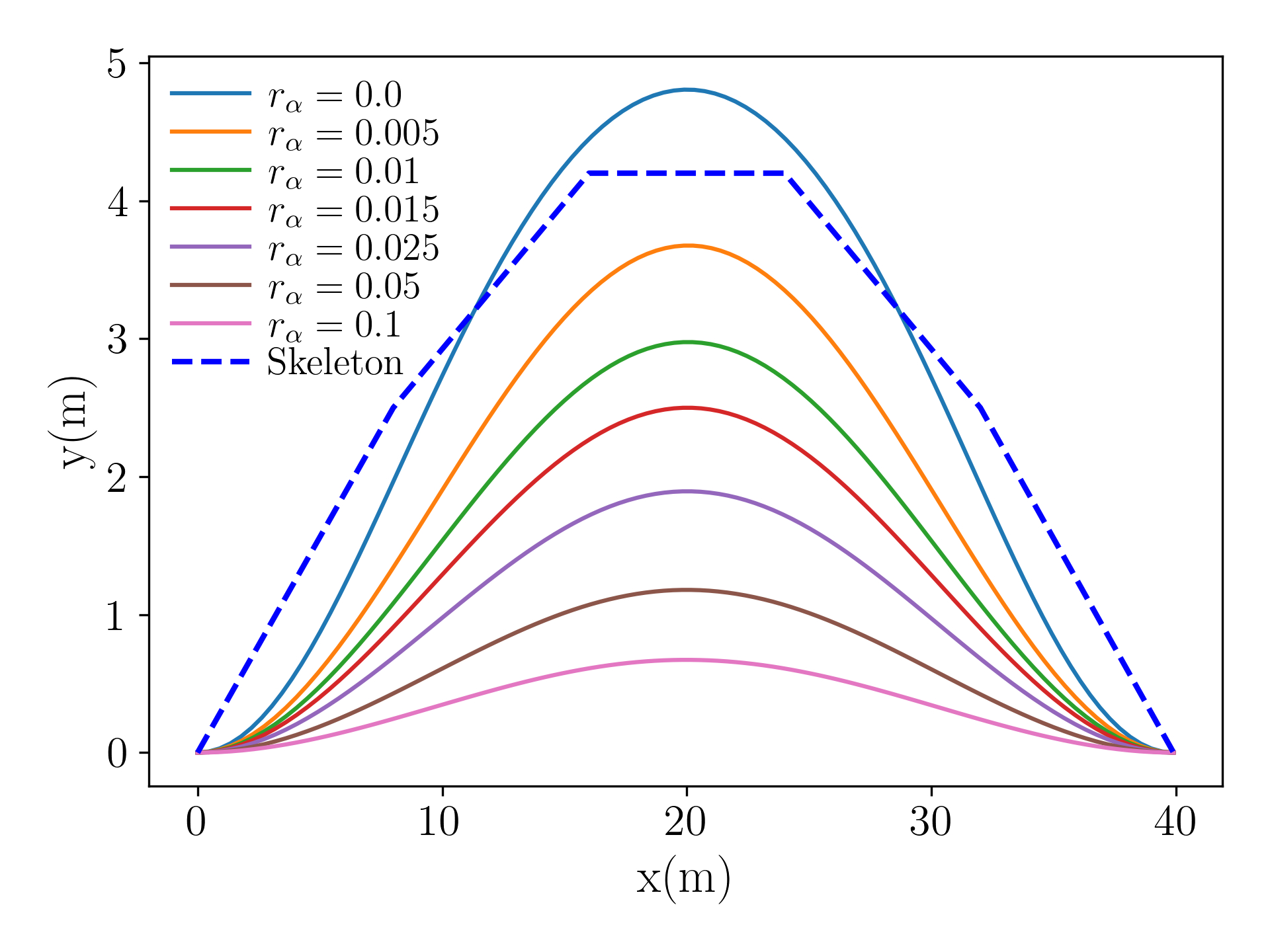}
  %% Use \caption command for figure caption and label.
  \caption{The overtaking trajectories generated by different fitting parameters. The blue dashed line represents the topological skeleton, while the realized overtaking trajectory is the fitted one. The larger the ratio of
    $\alpha_2/\alpha_1$, the greater the deviation of the overtaking trajectory from the topological skeleton.}\label{fig_diff_a}
\end{figure}

In this section, we evaluate how fitting parameters affect the trajectories. Given a topological skeleton and under the uncertainty of the given control inputs, the lower-layer planner evaluates different polynomial trajectories in parallel and selects the optimal one as the final overtaking trajectory. The polynomial trajectories are generated by (\ref{refine_traj}), where we fix $\alpha_1 = 1$ and vary the value of $\alpha_2$ 0 to 1.The fitting parameter scale factor  $r_\alpha = \frac{\alpha_2}{\alpha_1}$. A smaller $r_\alpha$ indicates that the fitted trajectory is closer to the topological skeleton. As shown in Figure \ref{fig_diff_a}, the shapes of the fitting curves for these different values are displayed. \revmwl{Since the topological skeleton is inherently collision-free by construction, the closer the trajectory adheres to the skeleton, the lower the collision risk, albeit at the expense of reduced smoothness.} Conversely, larger values of $r_\alpha$ result in higher smoothness for the trajectory, but they also lead to a greater deviation from the topological skeleton, which may increase the risk of collisions.
Here are descriptions of several key metrics used to evaluate the trajectory:
\begin{itemize}
  \item $\bm{len}$\label{len}: the length of trajectory
  \item $\bm{T}$\label{T}: the time to reach the trajectory endpoint.
  \item $\bm{d_{o,min}}$\label{d_omin}: the minimum distance between the trajectory and obstacles.
  \item $\bm{R_o}$\label{R_o}: the portion of the trajectory where the distance to obstacles is less than $r_{th}$ relative to the entire trajectory.
  \item $\bm{J_s}$\label{J_s}: the smoothness of the trajectory.
  \item $\bm{J_{\dot{\delta}}}$\label{J_delta}: the rate of change of the front wheel steering angle as the vehicle follows the trajectory.
\end{itemize}
Among the aforementioned metrics, $len$ and $T$ represent the efficiency of trajectory execution. $d_{o,min}$ and $R_o$ represent the safety of the trajectory. And $J_s$ and $J_{\dot{\delta}}$ represent the stability of the control inputs for the trajectory.
We define an operator $IP(\cdot)$ that represents the percentage improvement of the current value relative to a reference value. For example, $IP(T) = \frac{T_r-T_c}{T_r}$,  $IP(T)$ is the proportion by which the completion time $T_c$ of the current trajectory is reduced compared to the completion time $T_r$ of the reference trajectory.

The table \ref{tab:parameter of poly} indicates that as the value of $r_\alpha$ increases, the length of the generated trajectory decreases. In the case of $d_{o,min}$, it is evident that as $r_\alpha$ increases, the distance between the trajectory and the obstacle tends to decrease. However, this relationship is not strictly monotonic, as it is influenced by the motion trajectory of vehicles. For instance, when $r_\alpha = 0.005$, the minimum distance between the autonomous vehicle and another vehicle is 0.439 m. When $r_\alpha$ exceeds 0.015, the trajectory collides with the other vehicle. According to $R_o$, when $r_\alpha$ is less than 0.05, the minimum distance between the autonomous vehicle and the other vehicle exceeds $r_{th}$. However, when $r_\alpha$ surpasses 0.025, the proportion of the trajectory length that falls below $r_{th}$ gradually increases with the rising value of $r_\alpha$, eventually leading to collisions. The data from $d_{o,min}$ and $R_o$ suggest that when $r_\alpha$ reaches a certain threshold, the deviation of the trajectory from the reference path increases, thereby raising the risk of collision with the other vehicle. This threshold is variable, depending on the driving states of both vehicles and the road environment. To tackle this issue, the lower-layer planner employing the \hyperref[RPTG]{RPTG} method tests various $r_\alpha$ values and generates trajectories in parallel for safety assessments. This approach enables the automatic selection of $r_\alpha$ values that ensure trajectory safety in diverse scenarios.

A smaller $r_\alpha$ indicates that the fitted trajectory is closer to the reference path, resulting in greater safety. However, a lower $r_\alpha$ may also lead to reduced smoothness of the fitted trajectory. With $r_\alpha = 0$, the smoothness $J_s$ of the trajectory reaches 567.427 $m^2/s^5$, and as $r_\alpha$ increases, the smoothness of the trajectory improves rapidly. When $r_\alpha$ exceeds 0.05, there is more than a 90\% reduction in smoothness compared to $r_\alpha = 0$. Similarly, as $r_\alpha$ increases, the rate of change of the front wheel angle gradually decreases, indicating improved control stability of the trajectory. When $r_\alpha = 0.1$, the rate of change of the front wheel angle decreases by over 80\% compared to $r_\alpha = 0$. In summary, the data demonstrate that a smaller $r_\alpha$ results in higher safety for the trajectory, while a larger $r_\alpha$ enhances smoothness and control stability. Considering these factors together can help identify an optimal trajectory, such as the one corresponding to $r_\alpha = 0.015$ in the scenario.

\begin{table}[htbp]
	\caption{pure pursiut controller parameter settings}
	\centering
        \footnotesize
        \setlength{\tabcolsep}{10pt}
		\begin{tabular}{lll}
			\toprule
			Parameter & Description                     & Value        \\
			\midrule
			$L$       & Look-ahead distance coefficient & $0.1$        \\
			$L_{fc}$  & Look-ahead distance             & $0.8\,m/s^2$ \\
			$K_p$     & Speed P controller coefficient  & 5            \\
			$n$       & the horizon of controller       & 50           \\
			\bottomrule
		\end{tabular}

	\label{tab:pp_setting}
\end{table}

\begin{figure*}[!t]
  \centering

  % First row: image 1 and image 2
  \begin{minipage}[t]{0.49\textwidth}
    \centering
    \includegraphics[width=\textwidth]{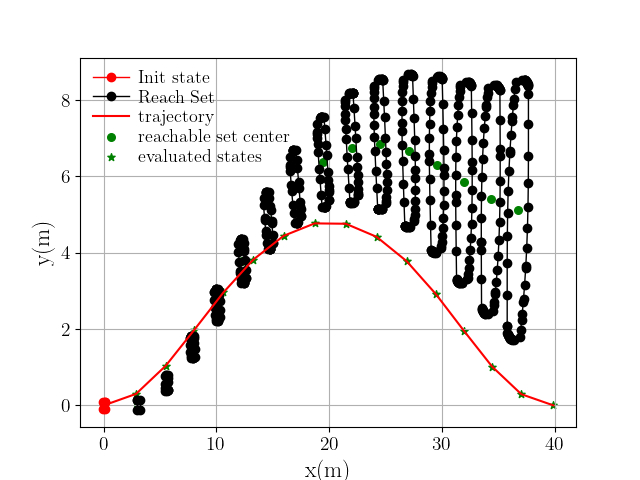}
    \subcaption{The reachable set corresponding to $r_\alpha=0.0$}
    \label{fig_reachset1}
  \end{minipage}
  \hfill
  \begin{minipage}[t]{0.49\textwidth}
    \centering
    \includegraphics[width=\textwidth]{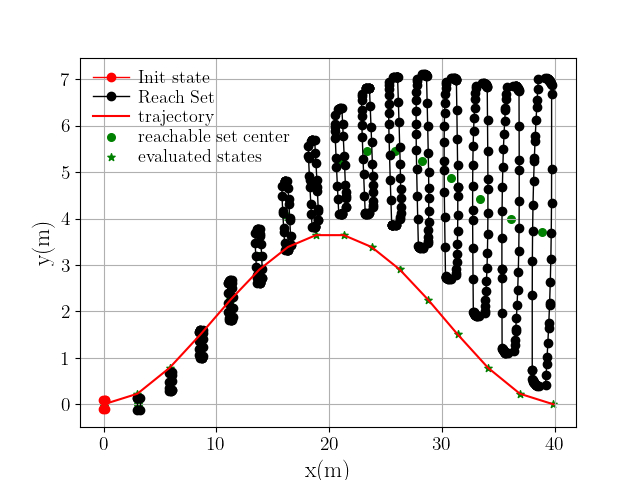}
    \subcaption{The reachable set corresponding to $r_\alpha=0.005$}
    \label{fig_reachset2}
  \end{minipage}

  \vspace{0.1cm} % Add some vertical space between rows

  % Second row: image 3 and image 4
  \begin{minipage}[t]{0.49\textwidth}
    \centering
    \includegraphics[width=\textwidth]{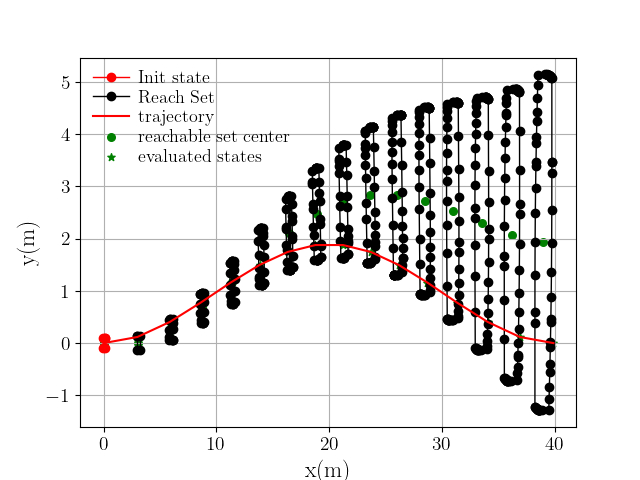}
    \subcaption{The reachable set corresponding to $r_\alpha=0.025$}
    \label{fig_reachset3}
  \end{minipage}
  \hfill
  \begin{minipage}[t]{0.49\textwidth}
    \centering
    \includegraphics[width=\textwidth]{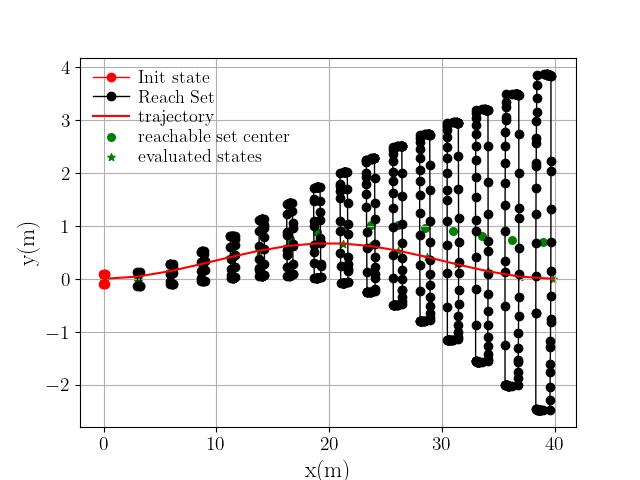}
    \subcaption{The reachable set corresponding to $r_\alpha=0.1$}
    \label{fig_reachset4}
  \end{minipage}
  \caption{\revmwl{Reachable sets corresponding to various fitting parameters.} The red trajectory represents the fitted generated trajectory. The green asterisks on the trajectory indicate equidistant time points along the path. The black dots forming a circle represent the reachable set corresponding to the green asterisks, and the green dots indicate the center of the corresponding reachable set. As $r_\alpha$ increases, the trajectory gets closer to the center of the reachable set, indicating its kinematic feasibility is relatively good. When the trajectory points fall outside the reachable set, it implies that the kinematic feasibility of the trajectory is relatively poor, such as in the case where $r_\alpha=0$.}
  \label{fig:four_reachable_set}
\end{figure*}

The step size $N_{RS}$ for analyzing the reachable set of trajectories is set to 15. As shown in Figure \ref{fig:four_reachable_set}, the larger the value of $r_\alpha$, the closer the overtaking trajectory is to the center of the reachable set, indicating higher feasibility of the trajectory. The metrics to describe the deviation of the trajectory from the reachable set are $J_s$, $J_v$, $J_{\theta}$ and their sum $J_{RS}$, which are shown in (\ref{rscost}). As shown in Table \ref{tab:parameter of poly}, when $r_\alpha = 0$, the trajectory exhibits poor kinematic feasibility because certain parts of the trajectory are not within the corresponding reachable set. \revmwl{Consequently, such trajectories are eliminated from the candidate set.} When $r_\alpha = 0.1$, the trajectory has a high level of kinematic feasibility, but collisions occur, thus it cannot be used as an overtaking trajectory either. When $r_\alpha = 0.015$, the trajectory is both controllable and collision-free. Based on the sum of $J_s$, $J_v$ and $J_\theta$ the trajectory with the smaller value is selected as the final overtaking trajectory.
\begin{table*}[htbp]
  \centering
  \caption{The various metrics of trajectories are generated by different fitting parameters. It indicates that the optimal trajectory is chosen based on a balance of kinematic feasibility, trajectory quality, and safety. The optimal fitting parameters are $r_\alpha= 0.015$.}
  \label{tab:parameter of poly}
  \begin{threeparttable}
    \footnotesize
    \setlength{\tabcolsep}{6pt}
      \begin{tabular}{ccccccccccccc}
        \toprule
        Parameter                        & \multirow{2}*{Col.\tnote{1}} & \multicolumn{4}{c}{Reachable metrics} & \multicolumn{7}{c}{Evaluation metrics}                                                                                                                                                                                                                                                         \\
        \cmidrule(lr){1-1}\cmidrule(lr){3-6}\cmidrule(lr){7-13}
        $r_\alpha$                            & ~                            & \hyperref[J_p]{$J_p$}                 & \hyperref[J_v]{$J_v$}                  & \hyperref[J_theta]{$J_\theta$} & \hyperref[rscost]{$J_{RS}$} & $len(m)$                 & $d_{o,min}(m)$          & $R_o$                          & $J_s(m^2/s^5)$            & $IP(J_s)$       & $J_{\dot{\delta}}(rad/s)$ & $IP(J_{\dot{\delta}})$ \\
        \midrule
        \textcolor{blue}{0.000}\tnote{2} & \textcolor{blue}{N}          & \textcolor{blue}{0.534}               & \textcolor{blue}{0.094}                & \textcolor{blue}{0.039}        & \textcolor{blue}{0.667}     & \textcolor{blue}{41.411} & \textcolor{blue}{0.386} & \textcolor{blue}{\textbf{0\%}} & \textcolor{blue}{567.427} & -               & \textcolor{blue}{10.711}  & -                      \\
        0.005                            & N                            & 0.338                                 & 0.030                                  & 0.028                          & 0.396                       & 40.814                   & 0.439                   & \textbf{0\%}                   & 217.634                   & 61.7\%          & 8.980                     & 16.2\%                 \\
        0.010                            & N                            & 0.276                                 & 0.027                                  & 0.023                          & 0.326                       & 40.535                   & \textbf{0.225}          & \textbf{0\%}                   & 143.803                   & 74.7\%          & 7.322                     & 31.6\%                 \\
        \rowcolor{gray!25}0.015          & N                            & 0.235                                 & 0.026                                  & 0.019                          & 0.280                       & 40.379                   & 0.331                   & \textbf{0\%}                   & 102.894                   & 81.9\%          & 6.169                     & 42.4\%                 \\
        0.025                            & Y                            & 0.183                                 & 0.025                                  & 0.015                          & 0.223                       & 40.218                   & 0.000                   & 23.6\%                         & 61.227                    & 89.2\%          & 4.687                     & 56.2\%                 \\
        0.050                            & Y                            & 0.124                                 & 0.025                                  & 0.009                          & 0.158                       & 40.158                   & 0.000                   & 44.2\%                         & 26.900                    & 95.3\%          & 2.925                     & 72.7\%                 \\
        0.100                            & Y                            & 0.085                                 & 0.025                                  & 0.005                          & \textbf{0.115}              & \textbf{40.075}          & 0.000                   & 78.5\%                         & \textbf{12.227}           & \textbf{97.9\%} & \textbf{1.669}            & \textbf{84.4\%}        \\
        \bottomrule
      \end{tabular}
    \begin{tablenotes}
      \footnotesize
      \item[1] Col. is used to indicate whether the trajectory collides with an obstacle. A value of N signifies no collision, while a value of \\Y indicates a collision has occurred. 
      \item[2] The data marked in \textcolor{blue}{blue} are the baseline for the column, used to calculate the corresponding $IP(\cdot)$ of the metric.
      \item[] \raisebox{0.5ex}{\colorbox{lightgray}{\rule{0ex}{0pt}}} \hspace{0.2em} represents various metrics of the selected overtaking trajectory. \textbf{Bold} indicates the best results under the same experimen-\\tal conditions. \revmwl{These conventions apply to all subsequent tables.}
    \end{tablenotes}
  \end{threeparttable}
\end{table*}
In this subsection, we explore how different combinations of fitting parameters affect the feasibility of trajectories. Given the distribution of obstacles and the current state of the ego vehicle, \revmwl{it is challenging to pre-determine which combination of fitting parameters will yield a high level of kinematic feasibility while remaining collision-free.} Thus, the parallel trajectory generation method based on the reachable set offers an efficient approach to identifying suitable fitting parameters. It systematically enumerates a range of fitting parameters and evaluates the associated trajectories for kinematic feasibility and safety in parallel. In this way, the planner can filter out the optimal overtaking trajectory.

\subsubsection{Evaluate the kinematic feasibility of the trajectory using a pure pursuit controller}
In this section, we evaluate the kinematic feasibility of the trajectories generated by \hyperref[RPTG]{RPTG} using a tracking controller.
We assess the kinematic feasibility of the trajectory by analyzing the performance of the pure pursuit controller \cite{ahn2021accurate}. Higher tracking performance indicates stronger kinematic feasibility for any given controller, while lower tracking performance suggests weaker kinematic feasibility.

Root Mean Square Error (RMSE) is used to measure the overall magnitude of an error across all time steps.
\begin{equation}
  E = \sqrt{\frac{1}{n} \sum_{i=1}^{n} e_i^2}\label{RMSE}
\end{equation}
where $E$ is the RMSE, $e_i$ is the tracking error (e.g., lateral, longitudinal, tracking or yaw error) at the $i-$th timestamp, and $n$ is the execution step of the controller. The relevant parameter settings of the pure pursuit controller are shown in Table \ref{tab:pp_setting}. We will analyze the tracking performance based on the following metrics.
\begin{itemize}
  \item $\bm{E_l}$\label{E_l}: the RMSE of lateral error, which measures the perpendicular distance from the vehicle to the reference trajectory.
  \item $\bm{E_p}$\label{E_p}: the RMSE of tracking deviation, which measures the Euclidean distance between the vehicle and the reference trajectory at each time step.
  \item $\bm{E_\theta}$\label{E_theta}: the RMSE of yaw error, which measures the difference in heading angle between the vehicle and the reference trajectory, normalized to lie within the range of $[-\pi,\pi]$
  \item $\bm{w_m}$\label{w_m}: mean yaw rate of trajectory. This quantifies the rate of change of the yaw of vehicle over time.
\end{itemize}

The data in Table \ref{tab:parameter of kinematic feasibility2} indicates that as $J_{RS}$ decreases, the tracking performance of the trajectory significantly improves. For instance, when $r_\alpha = 0.025$, the lateral tracking error is reduced by 64.5\% with $r_\alpha = 0.$, and the heading angle tracking error decreases by 64.5\%. When $r_\alpha = 0.1$, the reduction in lateral tracking error reaches 87.1\%, and the heading angle tracking error is reduced by 87.1\%. This demonstrates that as $J_{RS}$ decreases, the geometric tracking error during trajectory tracking becomes smaller, indicating greater trackability of the trajectory. Additionally, compared to $r_\alpha = 0$, the reduction in angular velocity during trajectory tracking is 87.3\% when $r_\alpha = 0.1$. This suggests that a lower $J_{RS}$ leads to smoother trajectory tracking control. Overall, the results show that smaller values of $J_{RS}$ correlate with better tracking performance by the controller, which includes reduced tracking errors and smoother control inputs. This reinforces the validity of using the reachable set to assess the kinematic feasibility of the trajectory, as it aligns with the observed trends in tracking performance.

\begin{table*}[!t]
  \centering
  \caption{The relationship between the kinematic feasibility of the trajectory and tracking performance. It indicates that the higher the kinematic feasibility, the better the trajectory tracking performance is.}
  \label{tab:parameter of kinematic feasibility2}
  \begin{threeparttable}
    \footnotesize
    \setlength{\tabcolsep}{10pt}
      \begin{tabular}{cccccccccc}
        \toprule
        Parameter                        & \multirow{2}*{\hyperref[rscost]{$J_{RS}$}} & \multicolumn{8}{c}{Tracking metrics}                                                                                                                                                       \\
        \cmidrule(lr){1-1}\cmidrule(lr){3-10}
        $r_\alpha$                            & ~                                          & $E_{l}(m)$                           & $IP(E_l)$       & $E_{p}$ ($m$)           & $IP(E_p)$       & $E_{\theta}(rad)$       & $IP(E_\theta)$  & $\omega_m(rad/s)$       & $IP(\omega_m)$  \\
        \midrule
        \textcolor{blue}{0.000}\tnote{1} & \textcolor{blue}{0.534}                    & \textcolor{blue}{0.124}              & -               & \textcolor{blue}{0.181} & -               & \textcolor{blue}{0.186} & -               & \textcolor{blue}{0.606} & -               \\
        0.005                            & 0.338                                      & 0.084                                & 32.3\%          & 0.173                   & 4.4\%           & 0.124                   & 33.3\%          & 0.404                   & 33.3\%          \\
        0.010                            & 0.276                                      & 0.068                                & 45.2\%          & 0.127                   & 29.8\%          & 0.102                   & 45.2\%          & 0.331                   & 45.4\%          \\
        0.015                            & 0.235                                      & 0.058                                & 53.2\%          & 0.108                   & 40.3\%          & 0.087                   & 53.2\%          & 0.281                   & 53.6\%          \\
        0.025                            & 0.183                                      & 0.044                                & 64.5\%          & 0.106                   & 41.4\%          & 0.066                   & 64.5\%          & 0.215                   & 64.5\%          \\
        0.050                            & 0.124                                      & 0.028                                & 77.4\%          & 0.097                   & 46.4\%          & 0.042                   & 77.4\%          & 0.135                   & 77.7\%          \\
        0.100                            & \textbf{0.085}                             & \textbf{0.016}                       & \textbf{87.1\%} & \textbf{0.084}          & \textbf{53.6\%} & \textbf{0.024}          & \textbf{87.1\%} & \textbf{0.077}          & \textbf{87.3\%} \\
        \bottomrule
      \end{tabular}
    \begin{tablenotes}
      \footnotesize
      \item[1] The data marked in \textcolor{blue}{blue} are the baseline for the column, used to calculate the corresponding $IP(\cdot)$ of the metric.
    \end{tablenotes}
  \end{threeparttable}
\end{table*}

\begin{figure}[!t]%% placement specifier
	\centering%% For centre alignment of image.
	\includegraphics[width=0.45\textwidth]{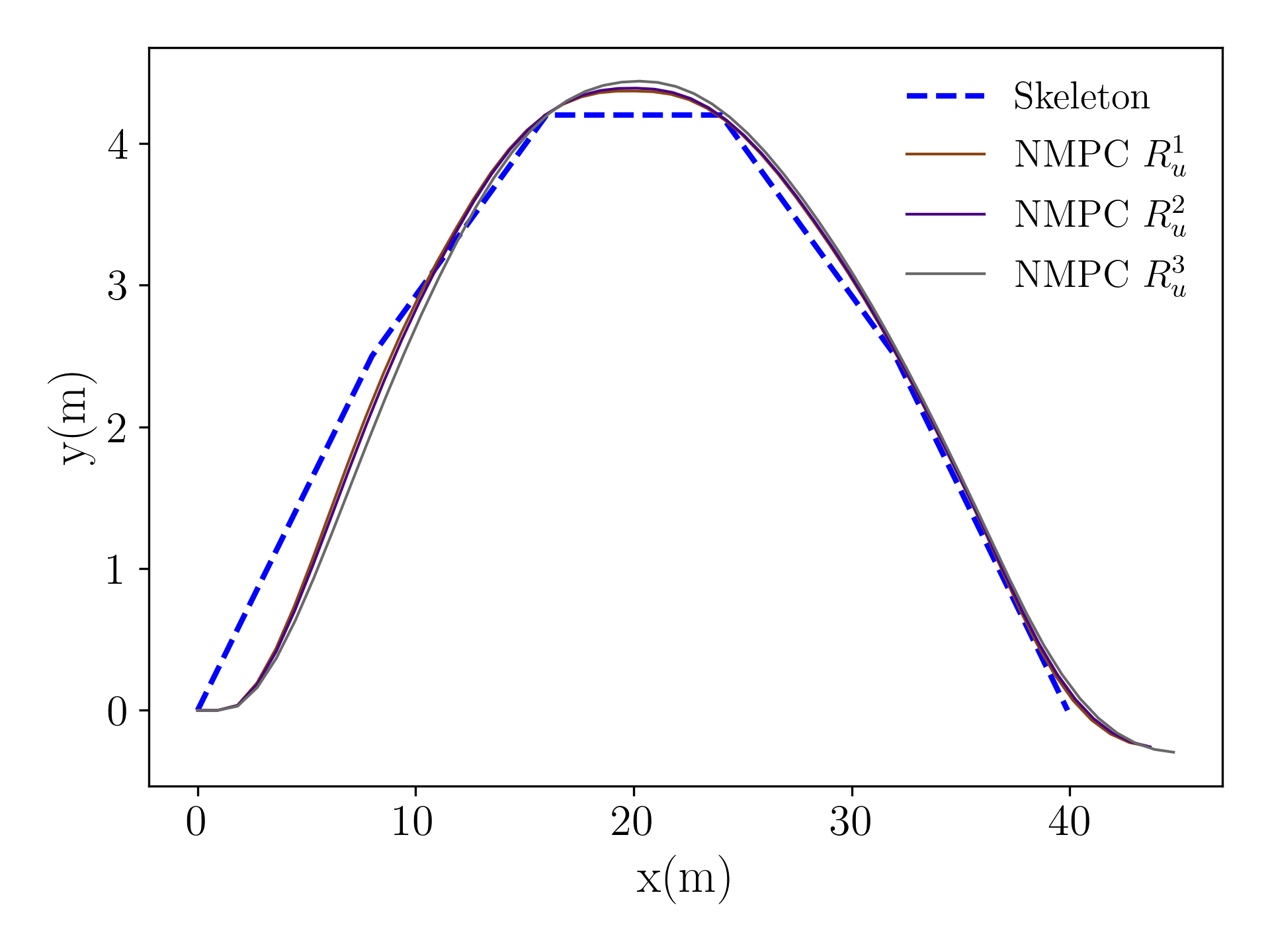}
	%% Use \caption command for figure caption and label.
	\caption{Comparison of the trajectory generated by NMPC. The position errors of NMPC trajectories show little variation across the different parameter settings compared to the topological skeleton.}\label{fig_diff_mpc_poly}
\end{figure}

\subsubsection{Comparison of our upper-layer trajectory planner and the NMPC planner}
\label{sec_compare_mpc}

In this section, we validate the effectiveness and efficiency of the proposed \hyperref[RPTG]{RPTG} method by comparing it with numerical methods. Given the same topological skeleton, the lower-layer planner samples both the proposed RPTG method and NMPC. 
%We use a nonlinear model predictive control (NMPC) method as the numerical approach. 
The basic form of the model predictive control is as follows:
\begin{equation}
  \begin{aligned}
    \min_{u} \quad    & \sum_{k=0}^{N-1} \left( \xi_k^\top \mathbf{R_s} \xi_k + u_k^\top \mathbf{R_u} u_k \right) + \xi_N^\top \mathbf{R_s} \xi_N \\
    \text{s.t.} \quad & \xi_{k+1} = f(\xi_k, u_k), \quad k = 0, 1, \dots, N-1,                                                                    \\
                      & \xi_0 = \xi_{\text{initial}},                                                                                             \\
                      & u_k \in \mathcal{U}, \quad \xi_k \in \mathcal{X}, \quad k = 0, 1, \dots, N.
  \end{aligned}
\end{equation}
where $\xi_{init}$ represents the initial values of the state variables, $\mathcal{X}$ represents the set of feasible state values, and $\mathcal{U}$ represents the set of feasible control values. Here, the horizon is chosen as $N = 15$. In the equations, $R_s$ represents the fitting cost coefficient to the skeleton, and $R_u$ represents the control input cost coefficient for the trajectory. We fix $R_s=diag([1.0,1.0,1.0,1.0])$ and provide three sets $R_u^1=diag([50.0,50.0]),R_u^2=diag([80.0,80.0]),R_u^3=diag([150.0,150.0])$ to evaluate the performance of NMPC. Testing has shown that a larger $R_u$ results in greater deviation between the trajectory and the skeleton, which can lead to collisions. Under the aforementioned settings, the trajectory generated by NMPC and the reference topological skeleton are shown in Figure \ref{fig_diff_mpc_poly}.

\begin{figure}[!t]%% placement specifier
	\centering%% For centre alignment of image.
	\includegraphics[width=0.45\textwidth]{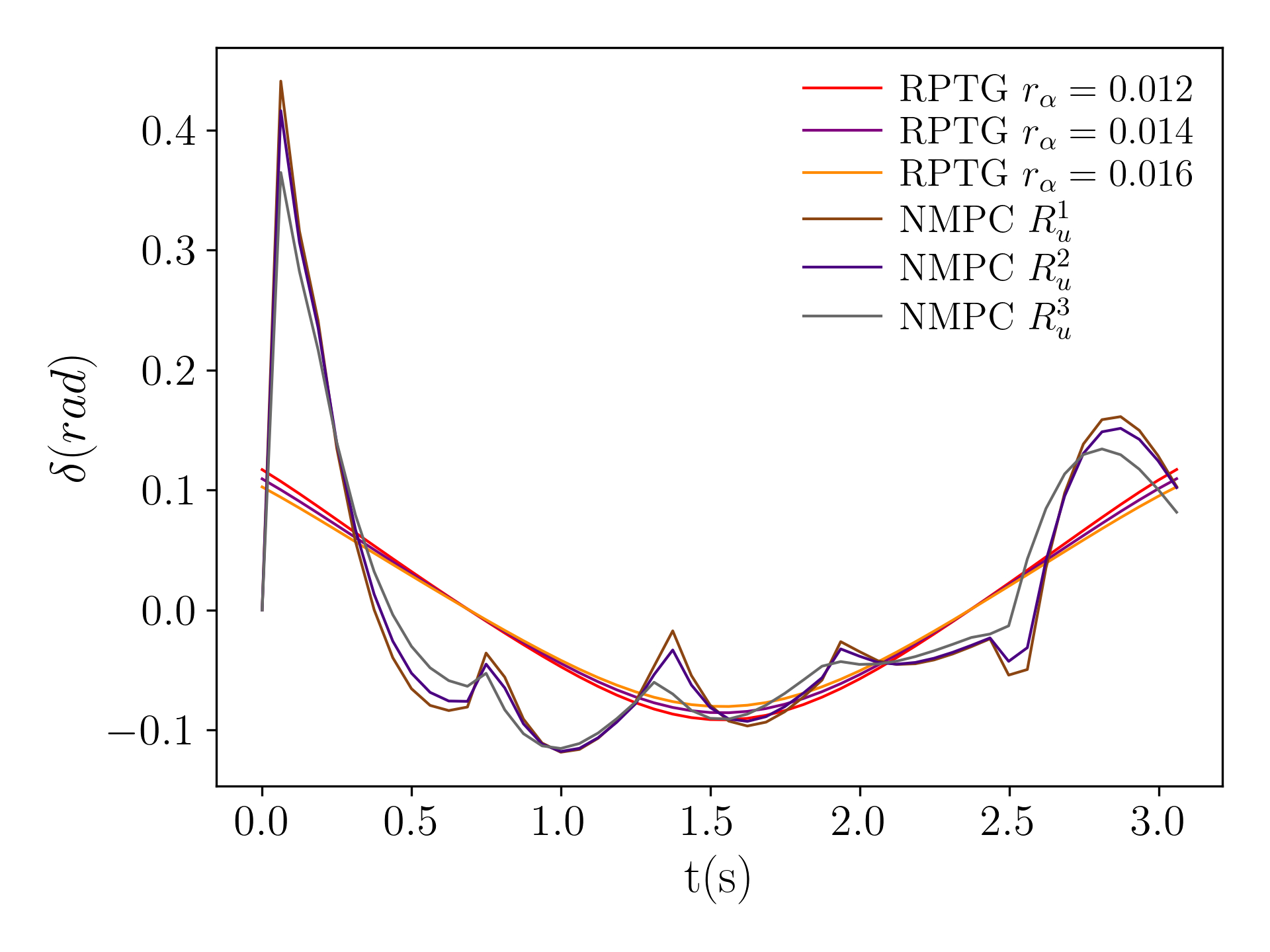}
	%% Use \caption command for figure caption and label.
	\caption{Comparison of the front wheel steering angles between trajectories generated by NMPC and RPTG.Compared to NMPC, the trajectory generated by RPTG has smaller front wheel steering inputs and relatively smaller changes in those inputs, indicating that the trajectory generated by RPTG has higher control stability.}\label{fig_diff_delta}
\end{figure}

As shown in the table \ref{tab:poly and mpc}, from the safety perspective, the NMPC method demonstrates better adherence to the topological reference skeleton, resulting in a greater minimum distance to obstacles compared to the RPTG method. Both methods maintain minimum distances that exceed the safe threshold distance $r_{th}$. Therefore, under the given parameter settings, the trajectories produced by both NMPC and RPTG meet the safety requirements. In terms of time efficiency, the RPTG method provides a significantly faster runtime, 61.6\% less than NMPC. Regarding the smoothness metric, RPTG achieves a 73.7\% improvement over the trajectories generated by NMPC. Figure \ref{fig_diff_delta} illustrates a comparison of the front wheel angles produced by both trajectory generation methods. The magnitude and variation of the front wheel angle in RPTG are smaller than those in NMPC, indicating that RPTG offers more excellent control stability. Overall, RPTG produces higher-quality trajectories compared to NMPC. Additionally, Since NMPC considers a nonlinear model, \revmwl{while RPTG decouples the vehicle model constraints from the optimization problem, the planning efficiency of the RPTG method is markedly superior to that of NMPC.}

\begin{figure*}[!t]
  \centering
  % First row: image 1 and image 2
  \begin{minipage}[t]{0.3\textwidth}
    \centering
    \includegraphics[width=\textwidth]{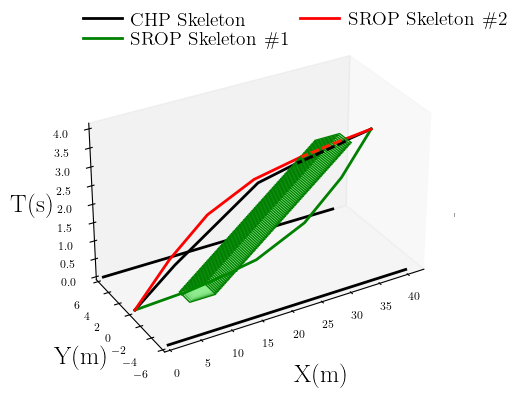}
    \subcaption{The schematic diagram of overtaking scenario $S1$ in the $s$-$l$-$t$ coordinate system.}
    \label{fig_road1}
  \end{minipage}
  \hfill
  \begin{minipage}[t]{0.3\textwidth}
    \centering
    \includegraphics[width=\textwidth]{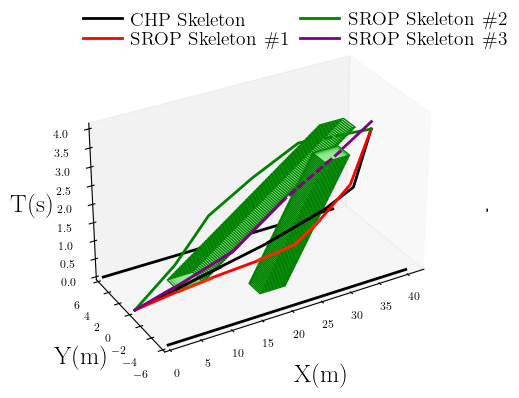}
    \subcaption{The schematic diagram of overtaking scenario $S2$ in the $s$-$l$-$t$ coordinate system.}
    \label{fig_road2}
  \end{minipage}
  \hfill
  \begin{minipage}[t]{0.3\textwidth}
    \centering
    \includegraphics[width=\textwidth]{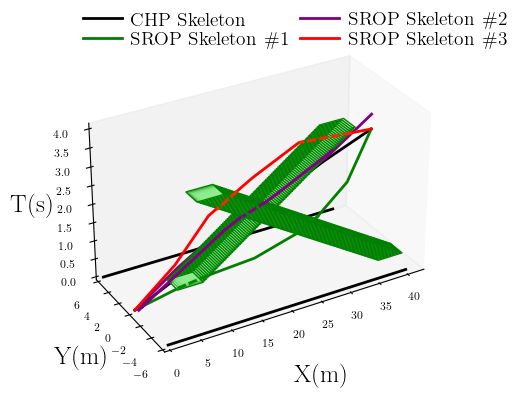}
    \subcaption{The schematic diagram of overtaking scenario $S3$ in the $s$-$l$-$t$ coordinate system.}
    \label{fig_road3}
  \end{minipage}

  \vspace{0.1cm} % Add some vertical space between rows

  % Second row: image 3 and image 4
  \begin{minipage}[t]{0.3\textwidth}
    \centering
    \includegraphics[width=\textwidth]{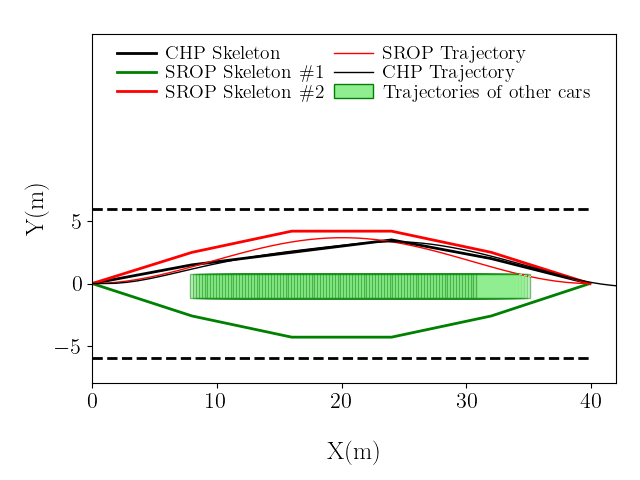}
    \subcaption{The top-down view of overtaking trajectory in scenario $S1$.}
    \label{fig_topo1}
  \end{minipage}
  \hfill
  \begin{minipage}[t]{0.3\textwidth}
    \centering
    \includegraphics[width=\textwidth]{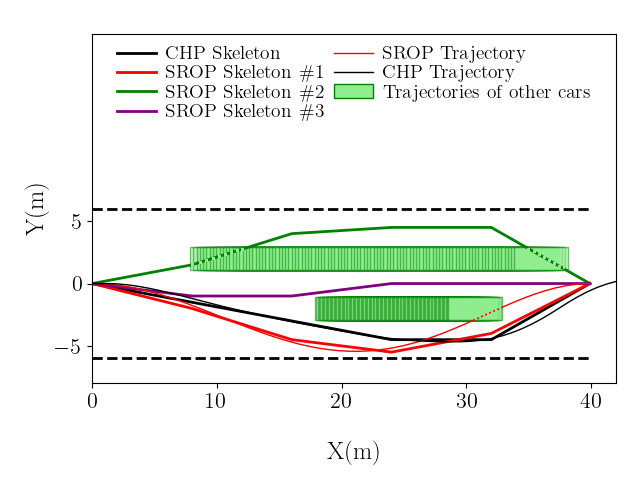}
    \subcaption{The top-down view of overtaking trajectory in scenario $S2$.}
    \label{fig_topo2}
  \end{minipage}
  \hfill
  \begin{minipage}[t]{0.3\textwidth}
    \centering
    \includegraphics[width=\textwidth]{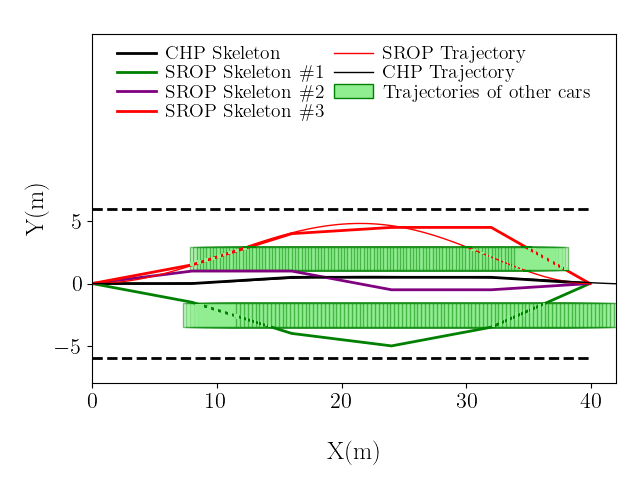}
    \subcaption{The top-down view of overtaking trajectory in scenario $S3$.}
    \label{fig_topo3}
  \end{minipage}
  \caption{(a),(b) and (c) are schematic diagrams of different overtaking scenarios in the $s$-$l$-$t$ coordinate system, and (d), (e), (f) are the top-down views of the corresponding scenarios. The green cylinders represent the trajectory of another vehicle. The red, green, and purple polylines are trajectory skeletons generated by SROP, each belonging to a different topological class. The black polyline represents the trajectory skeleton planned by DP. The red curve is the final overtaking trajectory fitted by SROP based on the corresponding skeleton, while the black curve represents the optimization of the DP planning result by NMPC. The dashed line in the figure indicates that the ego vehicle reaches this position ahead of the other vehicle. From the figure, it can be observed that in scenario $S1$, the trajectory planned by SROP and the trajectory planned by CHP belong to the same topological class, whereas in scenarios $S2$ and $S3$, they belong to different topological classes.}
  \label{fig:road_final_trajectory}
\end{figure*}

\begin{table*}[htbp]
  \centering
  \caption{\revmwl{Comparison of performance metrics between RPTG and NMPC planners.} The results show that in terms of runtime and trajectory quality, RPTG outperforms NMPC.}
  \label{tab:poly and mpc}
  \begin{threeparttable}
    \footnotesize
    \setlength{\tabcolsep}{10pt}
      \begin{tabular}{cccccccccc}
        \toprule
        \multirow{2}*{Mehthod}               & \multirow{2}*{Parameters} & \multirow{2}*{\hyperref[rscost]{$J_{RS}$}} & \multicolumn{6}{c}{Evaluation metrics}                                                                                                                                               \\
        \cmidrule(lr){4-10}
                                             & ~                         & ~                                          & $ T_r(ms)$                             & $IP(T_r)$       & $len(m)$                 & $d_{o,min}(m)$ & $v_{mean}$               & $J_{\dot{\delta}}(rad/s)$ & $IP(J_{\dot{\delta}})$ \\ \midrule

        \multirow{3}*{NMPC}
                                             & $R_u=R_u^1$               & -                                          & 192.311                                & -               & 43.911                   & 1.594          & 13.815                   & 27.621                    & -                      \\
                                             & $R_u=R_u^2$               & -                                          & 196.580                                & -               & 43.993                   & 1.564          & 13.839                   & 24.736                    & -                      \\
                                             & $R_u=R_u^3$               & -                                          & 187.262                                & -               & 45.063                   & 1.535          & 13.906                   & 20.167                    & -                      \\
        \textcolor{blue}{NMPC avg}\tnote{1}. & -                         & -                                          & \textcolor{blue}{192.051}              & -               & \textcolor{blue}{44.322} & -              & \textcolor{blue}{13.853} & \textcolor{blue}{24.175}  & -                      \\
        \midrule
        \multirow{3}*{RPTG}                  & $r_\alpha=0.012$               & 0.306                                      & 71.864                                 & 62.6\%          & 40.363                   & 0.486          & 13.228                   & 6.814                     & 71.8\%                 \\
                                             & $r_\alpha=0.014$               & 0.288                                      & 73.685                                 & 61.6\%          & 40.304                   & 0.364          & 13.209                   & 6.370                     & 73.7\%                 \\
                                             & $r_\alpha=0.016$               & 0.273                                      & 75.923                                 & 60.5\%          & 40.256                   & 0.353          & 13.193                   & 5.980                     & 75.3\%                 \\
        RPTG avg.                            & -                         & -                                          & \textbf{73.824}                        & \textbf{61.6\%} & \textbf{40.399}          & -              & \textbf{13.210}          & \textbf{6.358}            & \textbf{73.7\%}        \\

        \bottomrule
      \end{tabular}
    \begin{tablenotes}
      \footnotesize
      \item[1] The data marked in \textcolor{blue}{blue} are the baseline for the column, used to calculate the corresponding $IP(\cdot)$ of the metric.
    \end{tablenotes}
  \end{threeparttable}
\end{table*}

\subsubsection{Comparison of our \hyperref[SROP]{SROP} and other hierarchical planning method}
In this subsection, we validate the effectiveness and high efficiency of the proposed SROP method.
%, which is based on spatiotemporal topology and reachable set analysis. 
For comparison, we selected a \textbf{c}ommonly used \textbf{h}ierarchical trajectory \textbf{p}lanning method (CHP). In this method, the upper-layer planner uses a dynamic programming(DP) approach, while the proposed method employs the NMPC method described in Section \ref{sec_compare_mpc}. We compared these two planning methods across three overtaking scenarios. The upper-layer planner of SROP provides a set of multiple fitting parameters to determine the trajectory with optimal kinematic feasibility. The parameter $r_\alpha^\star$ \label{r_astar}represents the one from this set that minimizes $J_{RS}$ for the current skeleton path without causing any collisions.

\begin{itemize}
  \item S1 : there is one vehicle on the road moving in the same direction as the ego vehicle;
  \item S2 : there are two vehicles on the road moving in the same direction as the ego vehicle;
  \item S3 : there is one vehicle on the road moving in the same direction as the ego vehicle, and one vehicle moving in the opposite direction.
\end{itemize}

The comparison can be divided into two aspects. First, the proposed SROP method generates distinct topological path skeletons for different scenarios. % by considering the topological equivalence of the paths. 
In contrast, the dynamic programming-based approach does not take path topological equivalence into account and can only produce an initial path based on the cost function. As illustrated in Table \ref{tab:Hierarchical}, the optimal paths generated by CHP and SROP do not always belong to the same topological class. In Figure \ref{fig:road_final_trajectory}, the optimal trajectories of both CHP and SROP are within the same topological class in Scenario S1. However, in Scenario S2, the optimal trajectory generated by CHP falls within topological class \#3, while the optimal trajectory from SROP is in topological class \#2. Similarly, in Scenario S3, the optimal trajectory of CHP is in topological class \#2, whereas the trajectory produced by SROP is in topological class \#3.

In scenario S1, the optimal trajectory produced by the SROP method reduces the rate of change of the front wheel angle by 80\% compared to the CHP method. In scenario S2, this rate decreases by 72.4\% compared to CHP, while in scenario S3, the reduction is 71.6\%. These results suggest that, unlike CHP that focus solely on a single topological class, the SROP method is capable of accommodating initial solutions from various topological classes, which correspond to different local optimal solutions. By processing these initial solutions simultaneously and selecting the optimal trajectory, the overall quality of the solution can be significantly improved.

\begin{figure}[!t]%% placement specifier
  \centering%% For centre alignment of image.
  \includegraphics[width=0.45\textwidth]{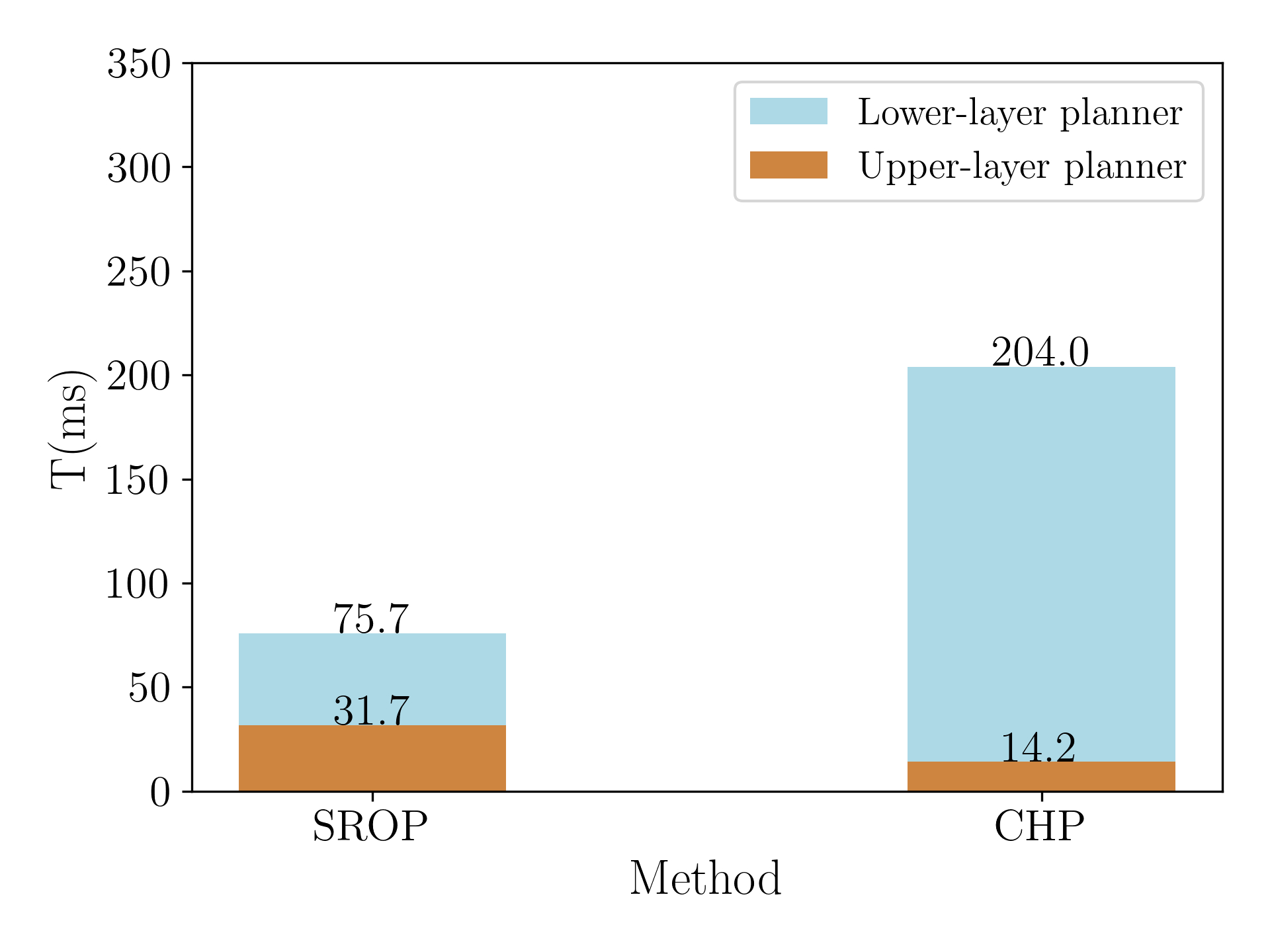}
  %% Use \caption command for figure caption and label.
  \caption{The time consumption comparison chart of SROP and CHP. The computation time of SROP decreased by 62.9\% compared to CHP, indicating that SROP has significantly higher planning efficiency.}\label{fig_diff_time}
\end{figure}

Secondly, in terms of efficiency, the dynamic programming method in CHP does take less time than the upper-layer planner of SROP. However, the trajectory optimization method using NMPC is significantly slower than the lower-layer planner of SROP. Overall, in scenarios S1, S2, and S3, the average computation time of the SROP method is reduced by 64.2\%, 58.7\%, and 62.4\%, respectively, compared to CHP. The SROP method achieved an overall reduction in computation time by 62.9\% compared to CHP, as illustrated in Figure \ref{fig_diff_time}. The quicker computation of the initial path in CHP is attributed to its dynamic programming method, which does not require consideration of topological equivalence. In contrast, the NMPC-based trajectory optimizer is time-consuming because it deals with a nonlinear model. The lower-layer planner of SROP efficiently addresses the polynomial fitting problem, allowing for rapid resolution. Moreover, extracting the reachable set of trajectories and assessing their kinematic feasibility consumes considerably less time than solving complex nonlinear optimization problems.

In summary, the numerical optimization-based hierarchical planning method proposed in this paper improves trajectory quality by considering multiple topological classes simultaneously, rather than focusing on a single one. Furthermore, by avoiding complicated nonlinear optimization models, this method significantly enhances efficiency.

\begin{table*}[htbp]
  \centering
  \caption{The various metrics of the trajectories planned by SROP and CHP. It shows that the planning results of the SROP method have better quality and higher efficiency.}
  \label{tab:Hierarchical}
  \begin{threeparttable}
    \footnotesize
    \setlength{\tabcolsep}{8pt}
      \begin{tabular}{ccccccccccc}
        \toprule
        \multirow{2}*{Scenario} & \multirow{2}*{Method}          & \multirow{2}*{Topo ID}\tnote{1} & \multirow{2}*{\hyperref[edgecost]{$J_\mathcal{E}$}} & \multicolumn{2}{c}{Reachable metrics} & \multicolumn{5}{c}{Evaluation metrics}                                                                                                                                                \\
        \cmidrule(lr){5-6} \cmidrule(lr){7-11}
                                & ~                              & ~                               & ~                                                   & \hyperref[rscost]{$J_{RS}$}           & \hyperref[r_astar] {$r_\alpha^\star$}       & $len(m)$                  & $T_r(ms)$\tnote{2}           & $IP(T_r)$                 & $J_{\dot{\delta}}(rad/s)$ & $IP(J_{\dot{\delta}})$    \\
        \midrule
        \multirow{3}*{S1}       & \multirow{2}*{SROP}            & \#1                             & 19.453                                              & 0.228                                 & 0.025                                  & 40.130                    & 33.1+40.5                    & 63.8\%                    & 4.814                     & 79.4\%                    \\
                                & ~                              & \cellcolor{gray!25}\#2          & \cellcolor{gray!25}19.667                           & \cellcolor{gray!25}0.223              & \cellcolor{gray!25}0.025               & \cellcolor{gray!25}40.118 & \cellcolor{gray!25}30.2+42.7 & \cellcolor{gray!25}64.2\% & \cellcolor{gray!25}4.687  & \cellcolor{gray!25}80.0\% \\
        \cmidrule(lr){2-11}
                                & \textcolor{blue}{CHP}\tnote{3} & \textcolor{blue}{2}             & -                                                   & -                                     & -                                      & \textcolor{blue}{43.059}  & \textcolor{blue}{15.2+188.3} & -                         & \textcolor{blue}{23.352}  & -                         \\
        \midrule
        \multirow{4}*{S2}       & \multirow{3}*{SROP}            & \#1                             & 19.942                                              & 0.365                                 & 0.010                                  & 40.661                    & 31.7+45.3                    & 61.4\%                    & 8.736                     & 57.0\%                    \\
                                & ~                              & \cellcolor{gray!25}\#2          & \cellcolor{gray!25}20.084                           & \cellcolor{gray!25}0.248              & \cellcolor{gray!25}0.020               & \cellcolor{gray!25}40.213 & \cellcolor{gray!25}34.2+48.3 & \cellcolor{gray!25}58.7\% & \cellcolor{gray!25}5.617  & \cellcolor{gray!25}72.4\% \\
                                & ~                              & \#3                             & 21.055                                              & 0.711                                 & 0.00                                   & 40.023                    & 32.4+41.5                    & 63.0\%                    & 7.216                     & 64.5\%                    \\
        \cmidrule(lr){2-11}
                                & \textcolor{blue}{CHP}          & \textcolor{blue}{\#3}           & -                                                   & -                                     & -                                      & \textcolor{blue}{42.148}  & \textcolor{blue}{13.7+185.8} & -                         & \textcolor{blue}{20.313}  & -                         \\
        \midrule
        \multirow{4}*{S3}       & \multirow{3}*{SROP}            & \#1                             & 17.784                                              & 0.327                                 & 0.010                                  & 40.500                    & 29.7+41.8                    & 65.8\%                    & 7.752                     & 58.6\%                    \\
                                & ~                              & \#2                             & 18.884                                              & 0.727                                 & 0.010                                  & 40.162                    & 30.9+44.8                    & 63.8\%                    & 9.273                     & 50.5\%                    \\
                                & ~                              & \cellcolor{gray!25}\#3          & \cellcolor{gray!25}19.357                           & \cellcolor{gray!25}0.238              & \cellcolor{gray!25}0.025               & \cellcolor{gray!25}40.181 & \cellcolor{gray!25}31.2+47.3 & \cellcolor{gray!25}62.4\% & \cellcolor{gray!25}5.326  & \cellcolor{gray!25}71.6\% \\
        \cmidrule(lr){2-11}
                                & \textcolor{blue}{CHP}          & \textcolor{blue}{\#2}           & -                                                   & -                                     & -                                      & \textcolor{blue}{42.138}  & \textcolor{blue}{13.6+195.4} & -                         & \textcolor{blue}{18.721}  & -                         \\
        \midrule
        avg.\tnote{4}           & -                              & -                               & -                                                   & -                                     & -                                      & -                         & -                            & \textbf{62.9\%}           & -                         & \textbf{66.8\%}           \\
        \bottomrule
      \end{tabular}
    \begin{tablenotes}
      \footnotesize
      \item[1] The same Topo ID indicates that the trajectories belong to the same topological class; otherwise, the trajectories belong to different topo-\\logical classes.
      \item[2] $T_r$ consists of two parts: the first part represents the runtime of the lower-layer planner, and the second part represents the runtime of the\\ upper-layer planner. 
      \item[3] The results of the CHP method in each scenario are used as the baseline for calculating the $IP(\cdot)$.
      \item[4] The average percentage improvement of SROP over CHP in terms of $T_r$ and $J_{\dot{\delta}}$ across all scenarios.
    \end{tablenotes}
  \end{threeparttable}
\end{table*}
\subsubsection{The improvement of \hyperref[STS]{STS} on trajectory quality and tracking performance}
This section assesses the enhancements in final trajectory performance by offering multiple initial paths from various topological classes. As illustrated in Table \ref{tab:Hierarchical}, several initial paths from different topological categories can be identified for various scenarios. In contrast, state-of-the-art methods usually present only a single initial solution.

Based on the cost value $J_\mathcal{E}$ of the topological skeleton, we can identify the path with Topo ID \#1 in scenarios S1, S2, and S3 for subsequent optimization. However, the final overtaking trajectories belong to different topological classes. This discrepancy arises because the distribution of obstacles means the solution space for the initial path may not always encompass the global optimum. Additionally, the initial path with the smallest $J_\mathcal{E}$ may be situated closer to obstacles.

During the optimization process conducted by the lower-layer planner, the value of $r_\alpha$ is increased to improve kinematic feasibility. However, this adjustment may also cause greater deformation of the optimized trajectory, leading to a higher risk of collisions. Conversely, initial paths with relatively larger values of $J_\mathcal{E}$ may allow for more deformation, which can ultimately result in better trajectory quality and enhanced kinematic feasibility.

\renewcommand{\arraystretch}{1.5}
\begin{table}[!t]
  \centering
  \caption{A comparison of tracking performance among trajectories belonging to different topological classes. The results show that this method can generate and find trajectories with better tracking performance from different topological classes.}
  \label{tab:ST_planner}
  \begin{threeparttable}
    {\huge
    \footnotesize
    \setlength{\tabcolsep}{1.5pt}
        \begin{tabular}{cccccccc}
          \toprule
          \multirow{2}*{Scenario} & \multirow{2}*{Topo ID}         & \multicolumn{4}{c}{Tracking metrics}                                                                                                                                           \\
          \cmidrule(lr){3-8}
                                  & ~                              & $E_{p}$ ($m$)                        & $IP(E_p)$                 & $E_{\theta}(rad)$        & $IP(E_\theta)$            & $\omega(rad/s)$          & $IP(\omega)$              \\
          \midrule
          \multirow{2}*{S1}       & \textcolor{blue}{\#1}\tnote{1} & \textcolor{blue}{0.106}              & -                         & \textcolor{blue}{0.068}  & -                         & \textcolor{blue}{0.221}  & -                         \\
                                  & \cellcolor{gray!25}\#2         & \cellcolor{gray!25}0.101             & \cellcolor{gray!25}4.7\%  & \cellcolor{gray!25}0.066 & \cellcolor{gray!25}2.9\%  & \cellcolor{gray!25}0.215 & \cellcolor{gray!25}2.7\%  \\
          \midrule
          \multirow{2}*{S2}       & \textcolor{blue}{\#1}          & \textcolor{blue}{0.164}              & -                         & \textcolor{blue}{0.120}  & -                         & \textcolor{blue}{0.386}  & -                         \\
                                  & \cellcolor{gray!25}\#2         & \cellcolor{gray!25}0.107             & \cellcolor{gray!25}34.8\% & \cellcolor{gray!25}0.079 & \cellcolor{gray!25}34.2\% & \cellcolor{gray!25}0.253 & \cellcolor{gray!25}32.7\% \\

          \midrule
          \multirow{2}*{S3}       & \textcolor{blue}{\#1}          & \textcolor{blue}{0.134}              & -                         & \textcolor{blue}{0.107}  & -                         & \textcolor{blue}{0.344}  & -                         \\

                                  & \cellcolor{gray!25}\#3         & \cellcolor{gray!25}0.108             & \cellcolor{gray!25}19.4\% & \cellcolor{gray!25}0.069 & \cellcolor{gray!25}35.5\% & \cellcolor{gray!25}0.253 & \cellcolor{gray!25}26.5\% \\
          \midrule
          avg\tnote{2}.           & -                              & -                                    & \textbf{19.6}\%           & -                        & \textbf{24.2\%}           & -                        & \textbf{20.6}\%           \\
          \bottomrule
        \end{tabular}}
    \begin{tablenotes}
      \footnotesize
      \item[1] The blue \textcolor{blue}{Topo ID} represents the trajectory with the smallest \hyperref[edgecost]{$J_\mathcal{E}$}\\ within that trajectory topological class. The resluts of the \textcolor{blue}{Topo ID} \\are the baseline for calculating the $IP(\cdot)$.
      \item[2] The average value is calculated by averaging the $IP(\cdot)$ across all\\ scenarios.
      \item[] \raisebox{0.5ex}{\colorbox{lightgray}{\rule{0ex}{0pt}}} \hspace{0.2em} represents various metrics of the selected overtaking trajectory.

    \end{tablenotes}
  \end{threeparttable}
\end{table}
As shown in Table \ref{tab:ST_planner}, the final overtaking trajectory exhibits notable improvements in both position tracking error $E_p$ and orientation angle error $E_\theta$ when compared to the trajectory with the smallest $J_\mathcal{E}$. Additionally, its average angular velocity $\omega_m$ is relatively lower. Across the three scenarios, the average reductions are 19.6\% for $E_p$, 24.2\% for $E_\theta$, and 20.6\% for $\omega_m$.

\begin{figure*}[!t]
	\centering
	
	% --- TOP ROW ---
	% First Subfigure (Top-Left)
	\subfloat[\label{fig:ros_overtake_1}]{%
		% Changed \textwidth to \linewidth
		\includegraphics[width=0.45\linewidth]{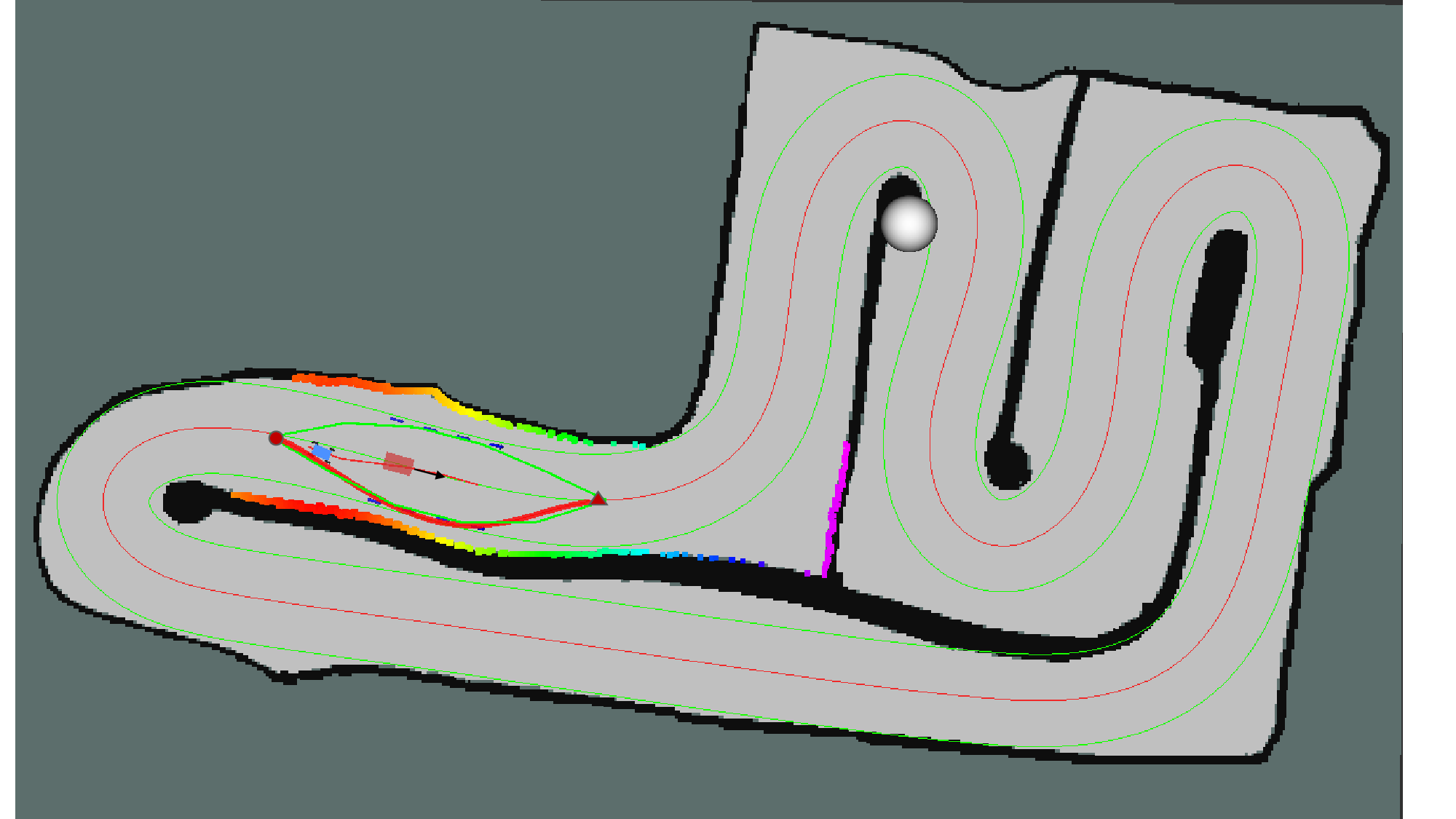}%
	}
	\hfill % Adds horizontal flexible space between top row images
	%  Second Subfigure (Top-Right)
	\subfloat[\label{fig:ros_overtake_3}]{%
		\includegraphics[width=0.45\linewidth]{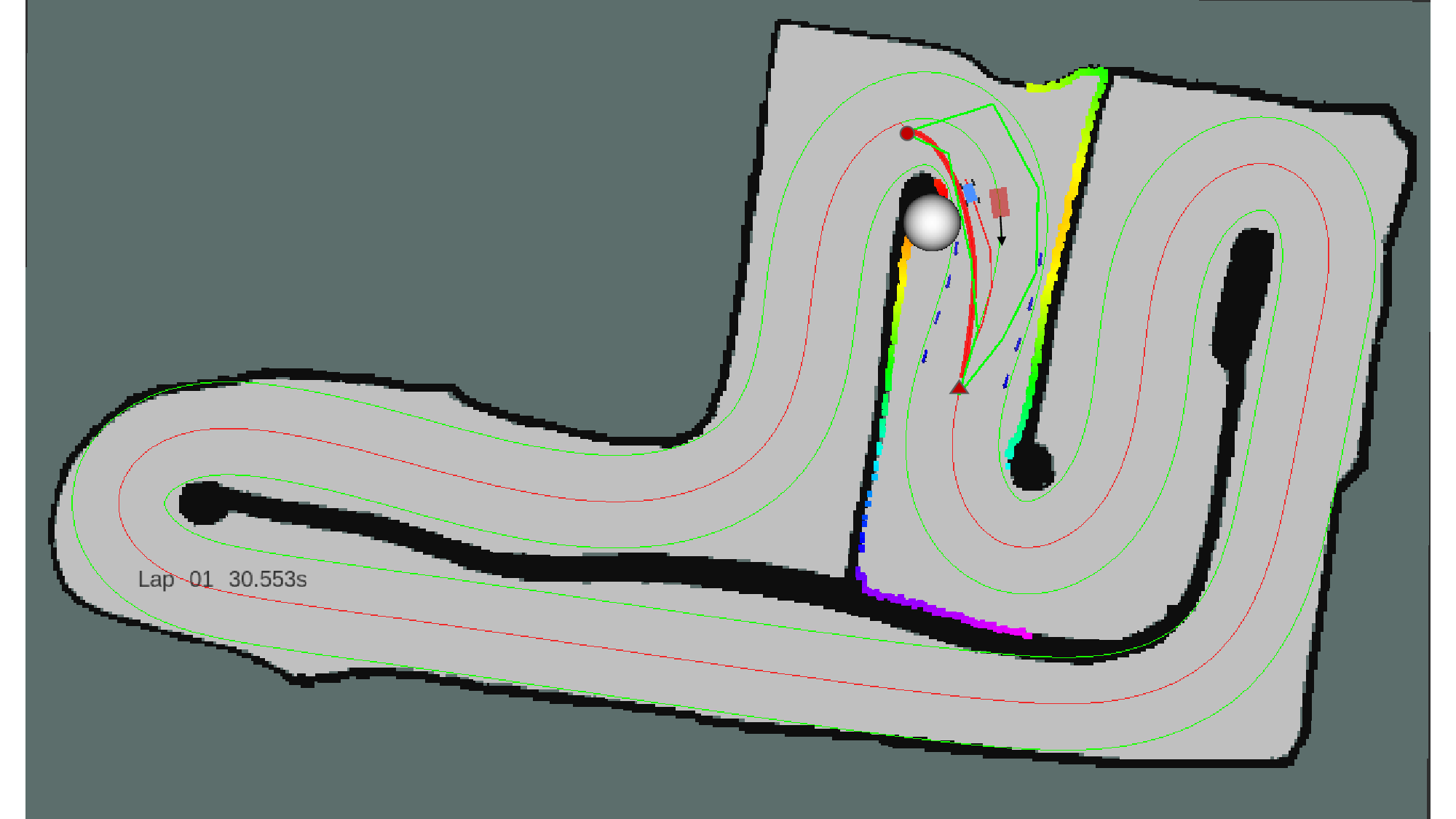}%
	}
	
	% Add vertical space between the rows
	\vspace{2ex}
	
	% --- BOTTOM ROW ---
	% Third Subfigure (Bottom-Left)
	\subfloat[\label{fig:ros_overtake_4}]{%
		\includegraphics[width=0.45\linewidth]{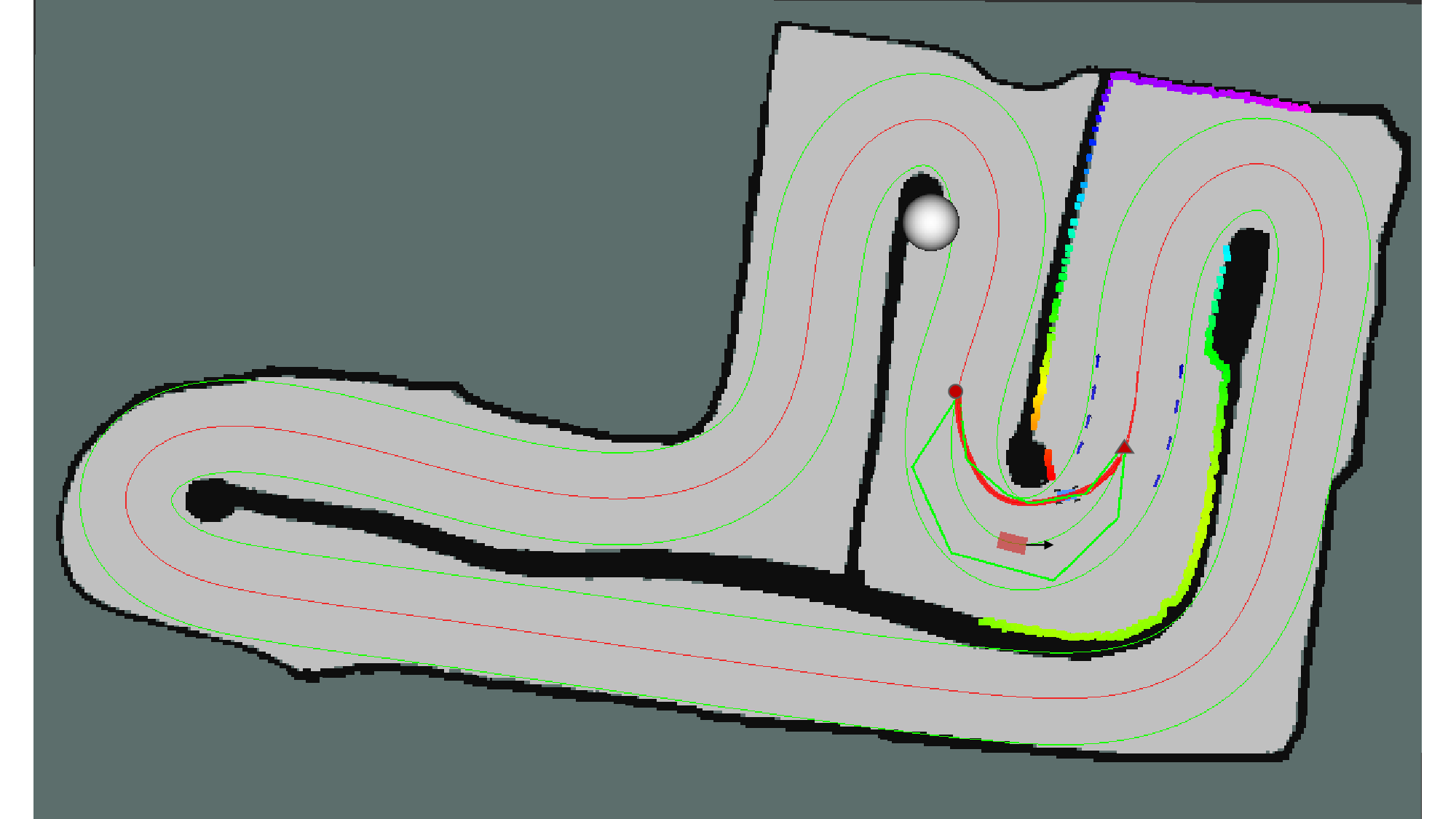}%
	}
	\hfill % FIXED: Added missing \hfill for the bottom row!
	% Fourth Subfigure (Bottom-Right)
	\subfloat[\label{fig:ros_overtake_2}]{%
		\includegraphics[width=0.45\linewidth]{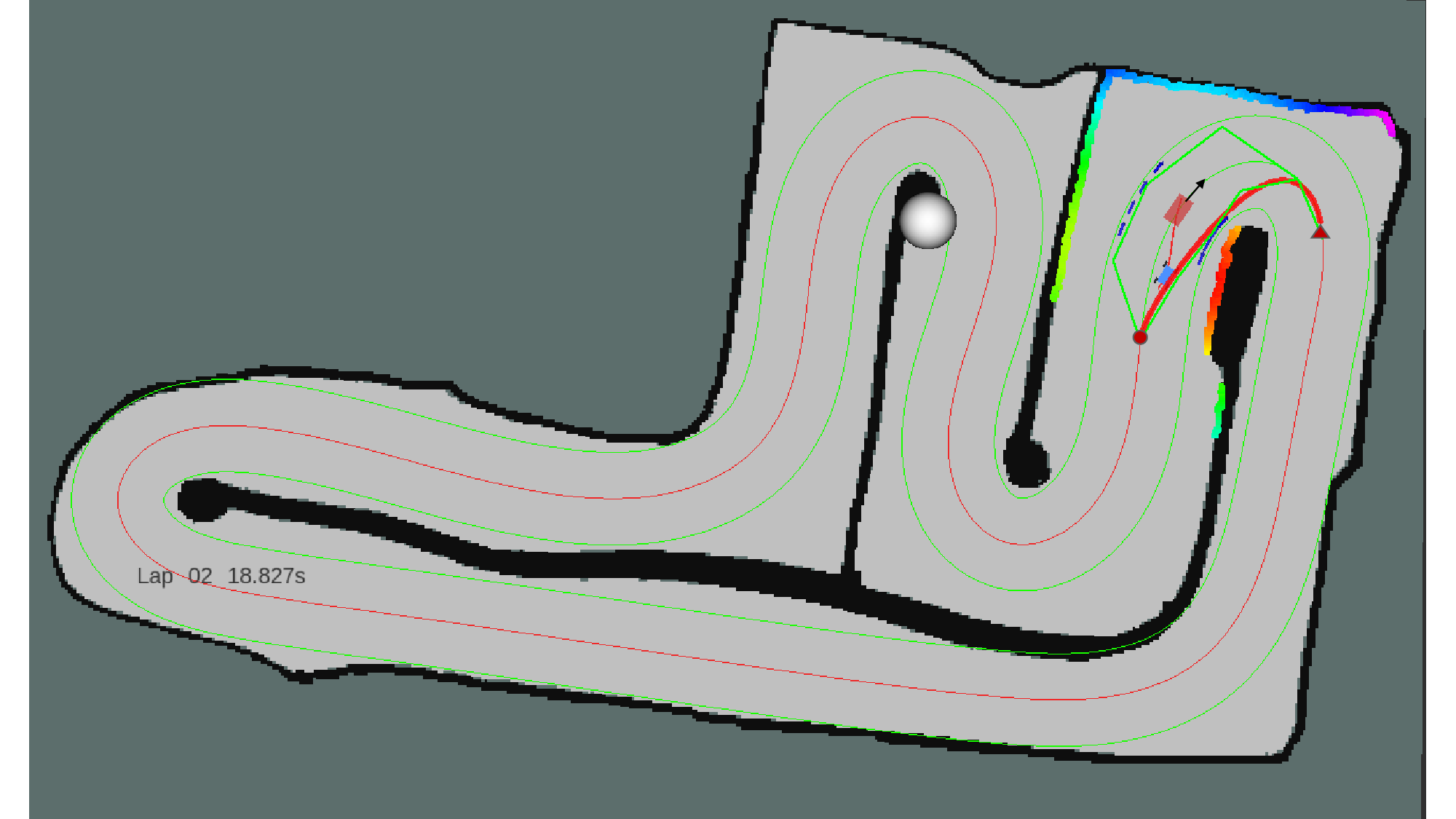}%
	}
	
	\caption{\revmwl{SROP-generated overtaking trajectories across various track segments in the F1TENTH simulator. Both vehicles travel clockwise, with the opponent (magenta rectangle) moving at 0.6 m/s. The green polyline and red curve denote the upper-layer skeleton and the optimal polynomial trajectory, respectively. Solid red circles and triangles mark the trajectory's start and end. A spherical indicator turns white to signal an active \textit{OVERTAKE} state.}}
	\label{fig:overtake_rviz}
\end{figure*}

The results indicate that the proposed method enhances trajectory quality compared to a single initial solution. This improvement arises from the ability of the method to generate initial paths from various topological classes for further optimization. Consequently, it avoids getting trapped in local optima, resulting in smoother trajectories and improved tracking performance.
\revmwl{\subsection{Integration with the F1TENTH Simulation Platform}
To demonstrate the practical applicability of the SROP method, it was integrated into the F1TENTH simulation platform \cite{o2020f1tenth}. Featuring accurate vehicle dynamics, realistic sensor pipelines, and robust low-level control at the limits of handling, this open-source environment serves as a reliable stepping stone, bridging theoretical trajectory planning with physical real-world applications. 
\subsubsection{Planning and Control Architecture in F1TENTH}
The ROS-based F1TENTH framework forms a complete autonomous software stack, sequentially integrating environment mapping, LiDAR-perception with Kalman Filter trajectory prediction, global racing-line planning, and precise trajectory execution via a Pure Pursuit controller.

A hierarchical Finite State Machine (FSM) governs the system's behavioral transitions among three primary states: \textit{GB\_TRACK}, \textit{OVERTAKE}, and \textit{TRAILING}. By default, the ego vehicle operates in the \textit{GB\_TRACK} state, following a global reference such as the optimal racing line or center line. The FSM initiates the overtaking sequence when a dynamic obstacle meets three specific conditions: a lower relative velocity, a sub-threshold longitudinal distance, and lateral obstruction of the ego vehicle's cruising path. Upon a successful SROP trajectory generation, the FSM formally commits to the \textit{OVERTAKE} state to execute the maneuver. Once the maneuver is completed and the vehicle merges back onto the global reference, it resumes the \textit{GB\_TRACK} state. Conversely, if the SROP planner fails—either (1) during execution due to a sudden obstacle cut-in, or (2) during the planning phase due to upper-layer topological unfeasibility or lower-layer collision risks—the FSM actively transitions to the \textit{TRAILING} state. In this state, the ego vehicle decelerates to follow the opponent safely until updated predictions re-trigger the overtaking sequence. Benefiting from the robust state management within the F1TENTH framework, this strict fallback mechanism ensures a high level of safety even in the event of overtaking planning failures.

\begin{figure*}[!t]
    \centering
    
    % --- Left Subfigure ---
    \subfloat[\revmwl{Evaluation scenario map and its segmentation.}\label{fig:map_arc}]{%
        \centering
        \includegraphics[width=0.48\linewidth]{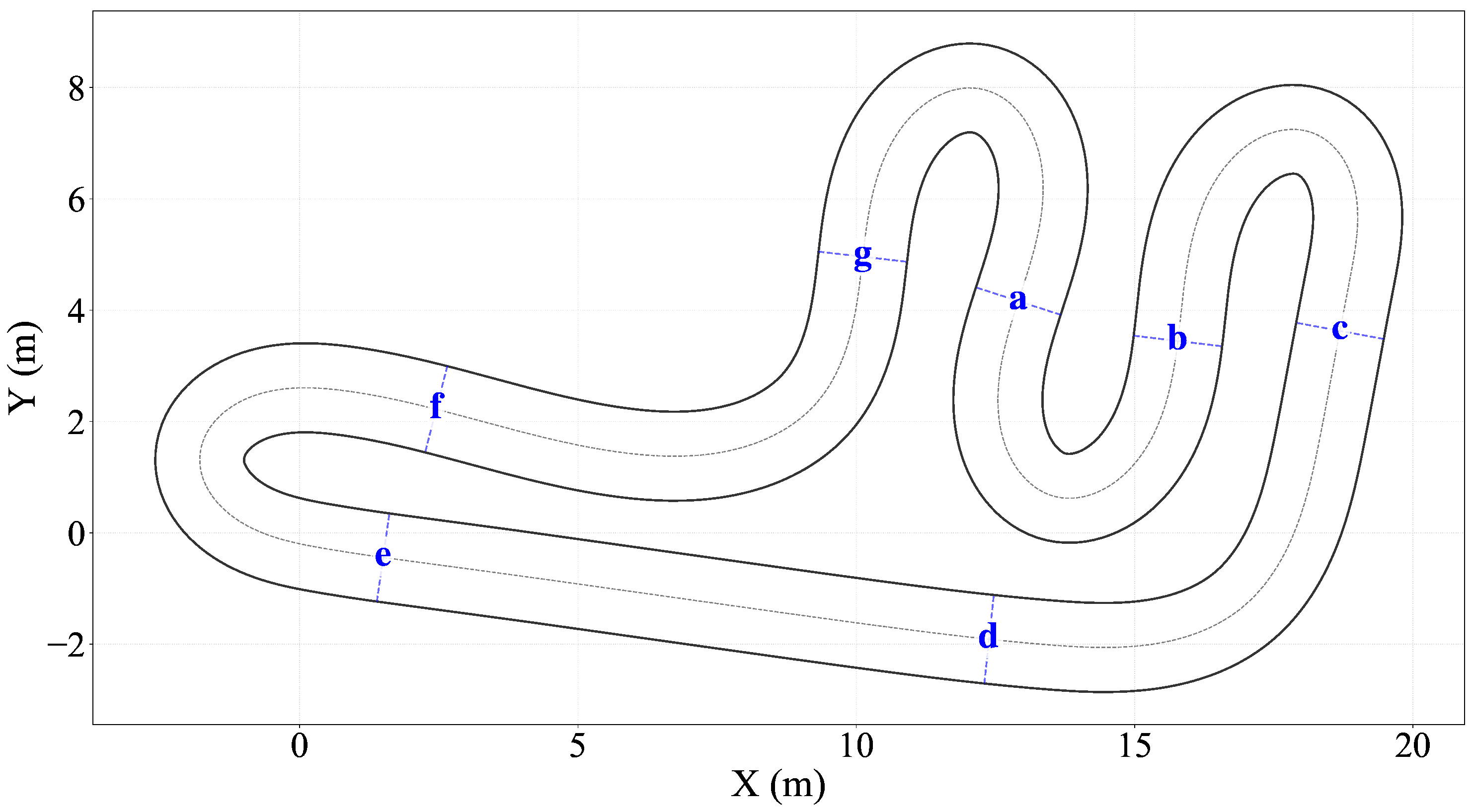}%
    }
    \hfill 
    % --- Right Subfigure ---
    \subfloat[\revmwl{Map curvature profile along the arc length.}\label{fig:map_curvature}]{%
        \centering
        \includegraphics[width=0.48\linewidth]{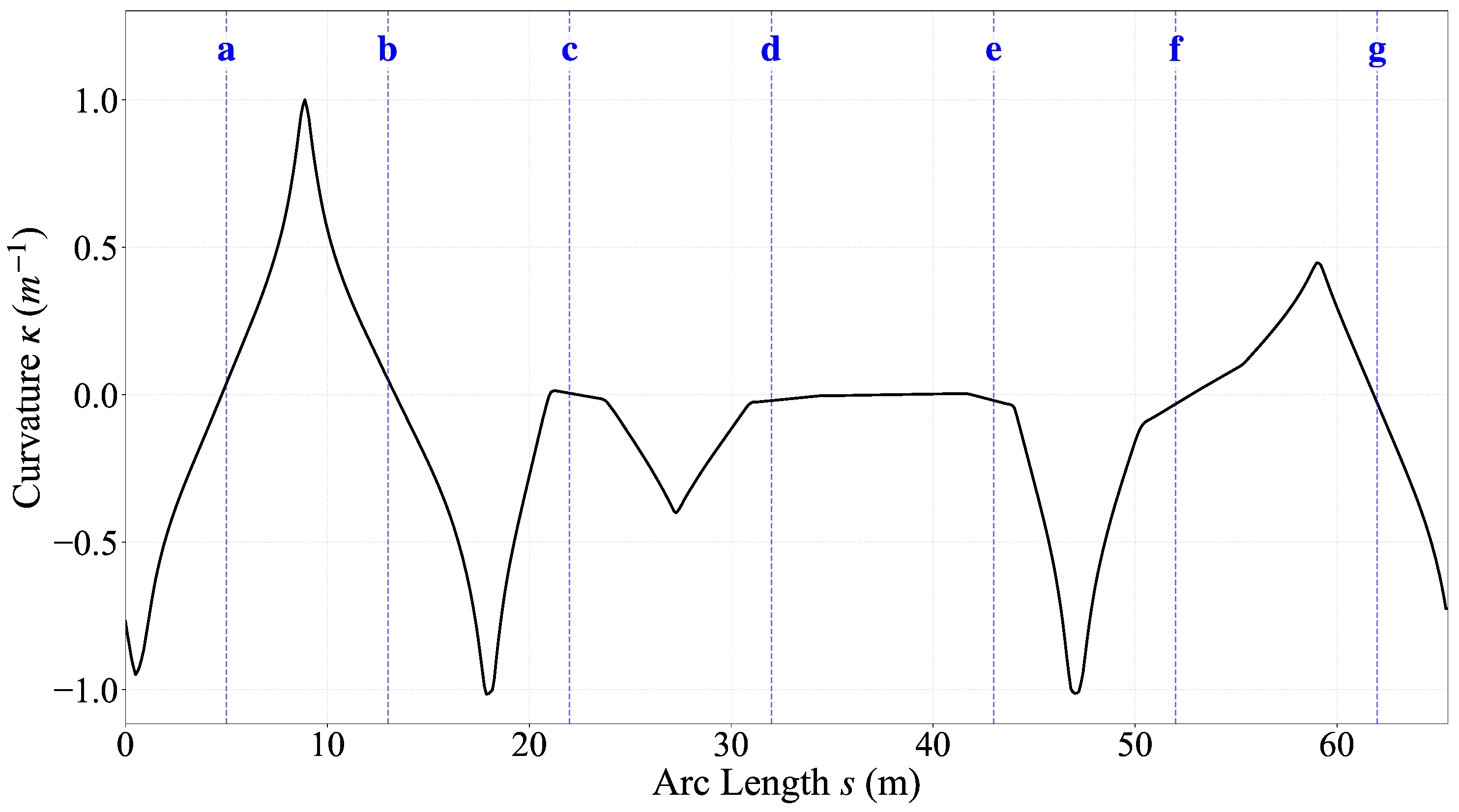}%
    }
    
    % --- MAIN CAPTION & LABEL ---
    \caption{\revmwl{Segmentation of the test map and its corresponding curvature profile.}}
    \label{fig:combined_map_analysis}
\end{figure*}

\subsubsection{Sensitivity Analysis in the F1TENTH Environment}
To quantitatively evaluate the SROP method's trajectory generation success rate under varying practical racing conditions, we conducted a comprehensive sensitivity analysis via a 100-lap continuous simulation on a closed-loop track. As shown in Figure \ref{fig:map_arc}, the track was divided into specific segments based on their average magnitude of ($|\bar{\kappa}|$). The corresponding curvature profile for each segment is illustrated in Figure \ref{fig:map_curvature}. The fitting parameter candidate set was $r_\alpha \in \{0.0, 0.001, 0.002, 0.005, 0.01, 0.015, 0.025, 0.05, 0.1\}$, and speed sensitivity was assessed by varying the opponent's constant speed ($v_{opp}$). To evaluate system performance, we recorded the overall success rate $R_{suc}$ (the ratio of successful overtakes $A_{suc}$ to attempts $A_{try}$), the average minimum obstacle clearance $\bar{d}_{o,min}$, and the mean computational time $\bar{T}_r$. Representative successful maneuvers are illustrated in Figure \ref{fig:overtake_rviz}.

The statistical results summarized in Table \ref{tab:overtaker_rate} reveal the distinct impacts of road curvature and opponent speeds on the overall overtaking success rate ($R_{suc}$).
\begin{itemize}\item \textbf{Sensitivity to Road Curvature:} The experimental data strongly aligns with vehicle kinematics. For instance, at an opponent speed of $v_{opp} = 1.5$ m/s, the nearly straight segment $de$ ($|\bar{\kappa}| = 0.005 \text{ m}^{-1}$) achieved a 100\% success rate. However, on the highly curved segment $ef$ ($|\bar{\kappa}| = 0.362 \text{ m}^{-1}$), $R_{suc}$ safely dropped to 40\%. Because navigating sharp curves consumes a significant portion of the lateral tire friction limit, the forward reachable set becomes highly restricted. SROP accurately captures this dynamically constrained envelope, actively rejecting unfeasible trajectories.
\item \textbf{Sensitivity to Relative Speeds:} The framework also demonstrates clear sensitivity to the relative speed advantage. Taking the sharpest curve $ab$ ($|\bar{\kappa}| = 0.418 \text{ m}^{-1}$) as an example, when $v_{opp}$ increased from 0.6 m/s to 1.8 m/s, $R_{suc}$ sharply decreased from 100\% to 25\%. A faster opponent demands a significantly prolonged spatio-temporal window to complete the maneuver, leaving little margin for maneuverability.
\end{itemize}
Overall, the proposed method achieves exceptional success rates under mild curvatures and conservative opponent speeds. This rate naturally decreases when confronting the kinematic limits of sharp curves and fast opponents. However, governed by a robust state machine, the system mitigates collision risks by seamlessly aborting unfeasible overtakes and reverting to a car-following mode. Ultimately, its successful F1TENTH integration validates its practical utility, demonstrating robust competitive performance and satisfying stringent real-time requirements with average execution times ($\bar{T}_r$) well below 100 ms.

\begin{table*}[htbp]
    \centering
    \caption{\revmwl{Overtaking performance metrics for varying opponent speeds and track segments in F1TENTH.}}
    \label{tab:overtaker_rate}
    
    % Increase row height slightly
    \renewcommand{\arraystretch}{1.2}
    % Reduce column separation slightly

    % Resize the table to fit the full page width (across both columns)
    \footnotesize
    \setlength{\tabcolsep}{2pt}
        % The table now has 19 columns in total
        \begin{tabular}{*{19}{c}}
            \toprule
            
            % First row of headers
            \multirow{2}{*}{Seg.} & 
            \multirow{2}{*}{$\left| \bar{\kappa} \right|(m^{-1})$} &
            \multicolumn{4}{c}{$v_{opp} = 0.6m/s$} & 
            \multicolumn{4}{c}{$v_{opp} = 1.0m/s$} & 
            \multicolumn{4}{c}{$v_{opp} = 1.5m/s$} & 
            \multicolumn{4}{c}{$v_{opp} = 1.8m/s$} &
            % Changed from \multirow to just putting the top part here
            $\bar{T_r}$ \\
            
            % Underlines for the velocity groups
            \cmidrule(lr){3-6} \cmidrule(lr){7-10} \cmidrule(lr){11-14} \cmidrule(lr){15-18}
            
            % Second row of headers
            & & $A_{try}$ & $A_{suc}$ & $R_{suc}$ & $\bar{d}_{o,min}(m)$ 
            & $A_{try}$ & $A_{suc}$ & $R_{suc}$ & $\bar{d}_{o,min}(m)$ 
            & $A_{try}$ & $A_{suc}$ & $R_{suc}$ & $\bar{d}_{o,min}(m)$ 
            & $A_{try}$ & $A_{suc}$ & $R_{suc}$ & $\bar{d}_{o,min}(m)$ & 
            % Put the unit part in the second row
            $(ms)$ \\
            \midrule
            
            % Placeholder data rows (with '-' at the end for T_r)
            $ab$  & 0.418 & 10 & 10 & 100\% & 0.192 & 8 & 7 & 87.5\% & 0.103 & 6 & 3 & 50\% & 0.067 & 4 & 1 & 25\% & 0.034 & 28.4+39.8 \\
            $bc$  & 0.361 & 6  & 6  & 100\% & 0.203 & 6 & 6 & 100\%  & 0.169 & 4 & 2 & 50\% & 0.081  & 5 & 1 & 20\% & 0.041 & 31.2+40.2 \\
            $cd$  & 0.150 & 17 & 17 & 100\% & 0.191 & 14& 14& 100\%  & 0.182 & 10& 9 & 90.0\%& 0.180 & 8 & 5 & 62.5\%& 0.081 & 27.2+38.5 \\
            $de$  & 0.005 & 19 & 19 & 100\% & 0.212 & 16& 16& 100\%  & 0.193 & 11& 11& 100\% & 0.217 & 10& 7 & 70\%  & 0.091 & 27.9+41.4 \\
            $ef$  & 0.362 & 7  & 7  & 100\% & 0.207 & 6 & 5 & 83.3\% & 0.145 & 5 & 2 & 40\%  & 0.052 & 3 & 0 & 0\%   & 0.0   & 26.6+40.8 \\
            $fg$  & 0.174 & 15 & 15 & 100\% & 0.187 & 11& 11& 100\%  & 0.201 & 9 & 8 & 88.9\%& 0.182 & 7 & 4 & 57.1\%& 0.071 & 30.4+41.2 \\
            $ga$  & 0.394 & 6  & 6  & 100\% & 0.171& 6 & 4 & 66.7\% & 0.093 & 5 & 2 & 40\%  & 0.055 & 3 & 1 & 33.3\%& 0.037 & 29.4+42.8 \\
            
            \bottomrule
        \end{tabular}%
    % End of resizebox
\end{table*}
}

\section{Conclusion}
\label{Conclusion}
\revmwl{This paper proposes a hierarchical overtaking trajectory planning framework to enhance computation efficiency and enforce kinematic feasibility in dynamic scenarios. The upper layer utilizes a spatio-temporal topological search to evaluate diverse initial paths, effectively avoiding local optima. The lower layer incorporates reachable set analysis into a parallel trajectory generation process, decoupling nonlinear vehicle constraints to efficiently produce kinematically feasible trajectories. Consequently, the proposed hierarchical planning method achieves significant improvements in both overall trajectory quality and time efficiency.

Numerical simulations demonstrate the superiority of the proposed method. By exploring multiple topological classes to avoid local optima, SROP reduces trajectory tracking error by 19.6\% and improves average smoothness by 20.6\%. Furthermore, evaluating kinematic feasibility via reachable sets decreases the tracking error of the final trajectory by 40.3\%. Overall, compared to state-of-the-art hierarchical planning methods, our framework improves trajectory smoothness by 66.8\% and reduces computation time by 62.9\%. The proposed method is further validated through seamless integration into the F1TENTH autonomous racing simulation platform. A rigorous 100-lap sensitivity analysis confirms that it maintains high overtaking success rates even in highly dynamic scenarios. These results strongly demonstrate the proposed method's practical utility, real-time efficiency, and dynamic safety in highly competitive environments.

Despite the promising results and practical utility demonstrated above, there remains room for further optimization. Specifically, the current framework's fixed discretization intervals inherently trade theoretical resolution completeness for computational efficiency. To address this, future work will explore adaptive sampling leveraging Vision-Language Models (VLMs) \cite{zhou2024vision}, \cite{pan2024vlp} to dynamically adjust discrete steps based on semantic scene understanding. Additionally, we plan to utilize offline neural network training to directly map trajectories to reachable sets, significantly accelerating online inference and overall planning in complex scenarios.}
\bibliographystyle{elsarticle-num-names}
\bibliography{SROP.bib}

\end{document}